\documentclass{article}

\PassOptionsToPackage{numbers, compress}{natbib}

    \usepackage[preprint]{neurips_2025}

\usepackage{configs/packages}
\usepackage{configs/commands}

\usepackage[utf8]{inputenc} 
\usepackage[T1]{fontenc}    
\usepackage{url}            
\usepackage{booktabs}       
\usepackage{amsfonts}       
\usepackage{graphicx}       
\usepackage{nicefrac}       

\usepackage{microtype}      
\usepackage{xcolor}         
\usepackage{amsmath}
\usepackage{textcomp}
\usepackage{colortbl}
\usepackage{xpatch}
\usepackage{tcolorbox}
\usepackage{mdframed}
\usepackage{enumitem}
\tcbuselibrary{skins}
\usepackage{multirow}
\usepackage{subcaption} 
\makeatletter
\xapptocmd{\NAT@bibsetnum}{\setlength{\leftmargin}{0pt}\setlength{\itemindent}{\labelwidth}\addtolength{\itemindent}{\labelsep}}{}{}
\makeatother

\title{Super Research: Answering Highly Complex Questions with Large Language Models through Super Deep and Super Wide Research }

\vspace{-2em}
\author{%
  Yubo Dong$^1$, Nianhao You$^1$, Yuxuan Hou$^1$, Zixun Sun$^1$, Yue Zhang$^1$ \\
  \textbf{Liang Zhang$^2$, Siyuan Zhao$^2$, Hehe Fan$^1$\footnotemark[1]} \\
  \\
  $^1$College of Computer Science and Technology \\
  $^2$Ant Group, Hangzhou, China \\
  \texttt{dongyubo@zju.edu.cn}, \quad  \texttt{hehefan@zju.edu.cn}
}
\begin{document}

\maketitle

\vspace{-2em}
\begin{abstract}
While Large Language Models (LLMs) have demonstrated proficiency in Deep Research or Wide Search, their capacity to solve highly complex questions—those requiring long-horizon planning, massive evidence gathering, and synthesis across heterogeneous sources—remains largely unexplored. We introduce \textbf{Super Research}, a task for complex autonomous research tasks that integrates (i) structured decomposition into a research plan, (ii) \textbf{super wide} retrieval for diverse perspectives, and (iii) \textbf{super deep} investigation to resolve uncertainties through iterative queries. 
To evaluate this capability, we curated a benchmark of 300 expert-written questions across diverse domains, each requiring up to 100+ retrieval steps and 1,000+ web pages to reconcile conflicting evidence. Super Research produces verifiable reports with fine-grained citations and intermediate artifacts (e.g., outlines and tables) to ensure traceable reasoning. Furthermore, we present a graph-anchored auditing protocol that evaluates Super Research along five dimensions: 
Coverage, Logical Consistency, Report Utility, Objectivity and Citation Health.
While super-complex questions may be infrequent in standard applications, Super Research serves as a critical ceiling evaluation and stress test for LLM capabilities. A model's proficiency within Super Research acts as a powerful proxy for its general research competence; success here suggests the robustness necessary to navigate nearly any subordinate research task. 
Leaderboard is available at: \url{https://cnsdqd-dyb.github.io/Super-Research-Benchmark/}
\end{abstract}


\begin{figure}[t]
    \centering
    \vspace{-3em}
    \includegraphics[width=0.7\linewidth]{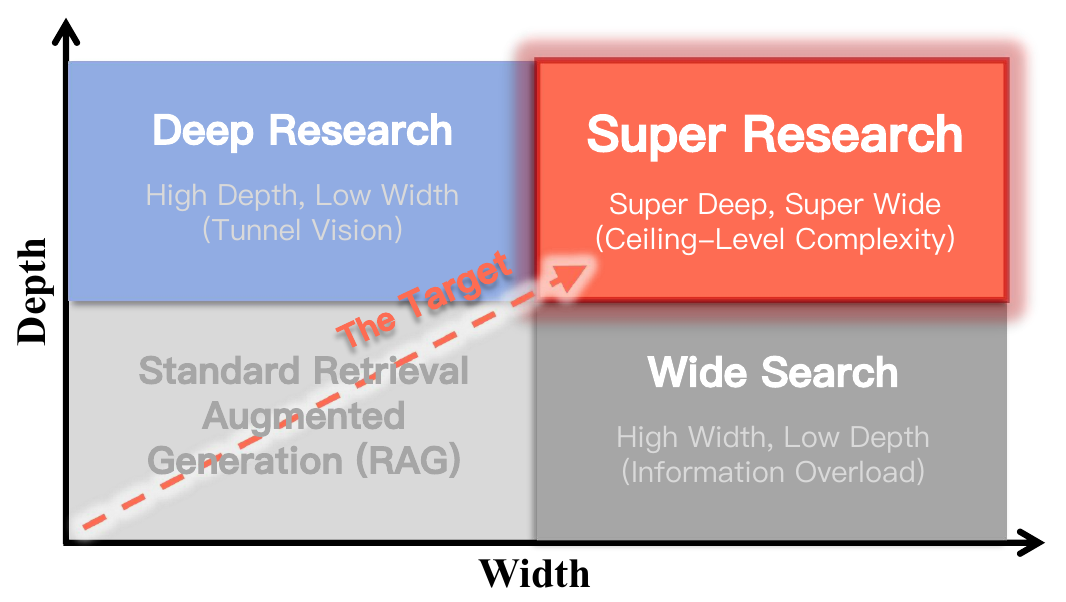} 
    \caption{Comparison of Super Research with standard Retrieval-Augmented Generation (RAG), Deep Research, and Wide Search. \textbf{(a) RAG} represents the baseline paradigm, which typically operates with limited depth and width. \textbf{(b) Deep Research} focuses on vertical exploration and recursive chains of evidence to resolve nuanced questions, though it often lacks horizontal width, leading to ``tunnel vision''. \textbf{(c) Wide Search} prioritizes horizontal data acquisition and large-scale coverage across diverse information nodes, but lacks synthetic depth, which can result in ``information overload''. \textbf{(d) Super Research} explicitly couples Super Deep investigation with Super Wide retrieval to address highly complex questions requiring long-horizon planning, 100+ retrieval steps, and the synthesis of 1,000+ web pages.  It can generate research reports of up to 50 pages, averaging around 100k  words per report.}
    \label{fig:intro}
\end{figure}

\section{Introduction}\label{sec:intro}

The evolution of Large Language Models (LLMs) has marked a paradigm shift in artificial intelligence, moving from simple statistical text completion to sophisticated reasoning and agentic problem-solving. Early iterations focused on scaling parameters to capture world knowledge within static weights. However, the recent shift toward Retrieval-Augmented Generation (RAG)~\cite{lewis2020retrieval} and agentic workflows~\cite{yao2022react, wu2024autogen} has empowered LLMs to interact with the external web, allowing them to overcome the limitations of their training cutoff and hallucination tendencies. Today, LLMs are no longer just repositories of information but are increasingly treated as ``reasoning engines'' capable of orchestrating complex sequences of tool use and information retrieval to answer non-trivial queries.

In tandem with these advancements, the concept of Deep Research~\cite{deepresearchbench, google2025deepresearch} has emerged as a specialized capability for LLMs. This approach prioritizes vertical exploration, where an agent iteratively follows a chain of evidence to resolve specific, nuanced questions. By employing ``Chain-of-Thought'' prompting~\cite{wei2022chain} and recursive search loops, Deep Research agents can dive into technical documentation, legal archives, or scientific papers to extract precise answers that are hidden beneath layers of text. The focus here is on the depth of inquiry, which ensures that a specific thread of reasoning is followed to its logical conclusion through rigorous verification and iterative refinement. 

Conversely, Wide Search~\cite{widesearch} (note that ``Wide Research'' is not yet established) has evolved to address the necessity for horizontal information coverage. This current paradigm prioritizes large-scale data acquisition, striving to capture an exhaustive array of information nodes and web sources to maximize the width of the raw dataset.
In contrast, we define Wide Research as a more sophisticated evolution that focuses on thematic width and perspective synthesis. Rather than mere data accumulation, Wide Research systems are designed to ``survey the landscape'', ensuring that no critical angle—be it economic, social, or technical—is overlooked. This is achieved through multi-query expansion and aspect-based retrieval, allowing the model to identify the multifaceted dimensions of a problem and curate a representative sample of evidence from across the digital ecosystem.

In this paper, we present Super Research, which is designed to push the boundaries of LLM capabilities in complex research. Super Research is dedicated to addressing super-complex questions such as \textit{``optimizing immunopharmacological mechanisms where T cell activation must be precisely balanced against tumor microenvironment immune escape pathways to minimize autoimmune risks''}, which lie far beyond the operational limits of current Deep Research or Wide Search paradigms. Super Research explicitly couples three core pillars: (i) Structured Decomposition, which breaks a monolithic query into a multi-layered research plan; (ii) Super Wide Retrieval, which explores the search space horizontally to ensure total coverage of diverse perspectives; and (iii) Super Deep Investigation, which utilizes iterative follow-up queries to resolve uncertainties and verify the reliability of individual data points. To rigorously test this framework, we introduce a new benchmark of 300 expert-written questions that require massive evidence gathering and synthesis. 

Compared to existing Deep Research or Wide Search that typically operate within a range of 10–20 retrieval iterations and approximately 100 web pages, Super Research is designed for a higher tier of complexity. 
Those truly ``super complex'' questions are found in professional intelligence, scientific discovery, or strategic planning, and usually demand long-horizon planning that can manage 100+ retrieval steps and 1,000+ web pages to synthesize massive evidence while maintaining both extreme width and extreme depth.  

However, evaluating Super Research is itself extremely challenging. Existing evaluation paradigms~\cite{gaia, deepresearchbench, wang2025liveresearchbench, hu2025step} are ill-suited for the scale and nuance of complex inquiry. 
\ding{172} \textbf{Inaccurate LLM-as-a-Judge:} Standard report-to-report comparisons~\cite{zheng2023judging} are prone to alignment errors and often fail to capture deep reasoning.
\ding{173} \textbf{Shallow Fact-Recall Metrics:} Current benchmarks~\cite{min2023factscore, wei2024long} prioritize ``atomic'' facts while neglecting the sophisticated synthesis and analytical depth required for complex questions.
\ding{174} \textbf{Costly Human Evaluation:} Expert reviews, and Elo-based systems~\cite{wan2025deepresearch} are rigorous but prohibitively slow, expensive, and difficult to scale.
\ding{175} \textbf{Neglected Ambiguity:} Most metrics fail to reward ``uncertainty expression'', penalizing models that correctly identify lack of consensus while rewarding false confidence. 

To bridge this gap, we formalize a structured evaluation protocol that assesses reports along five dimensions: Coverage Comprehension, Logical Consistency, Report Utility, Objectivity Score, and Citation Health.
Furthermore, we introduce an automated, graph-anchored auditing tool (See Appendix~\ref{app:web_tool} ). By projecting generated reports onto an expert-curated knowledge graph, this tool enables:
\ding{172} \textbf{Structured Claim Verification:} Mapping claims directly to verified data.
\ding{173} \textbf{Interpretable Error Tracing:} Pinpointing exactly where a reasoning chain breaks.
\ding{174} \textbf{Reproducible Scoring:} Ensuring fair and consistent evaluation at scale.

\begin{figure*}[t]
    \centering
    \includegraphics[width=0.99\linewidth]{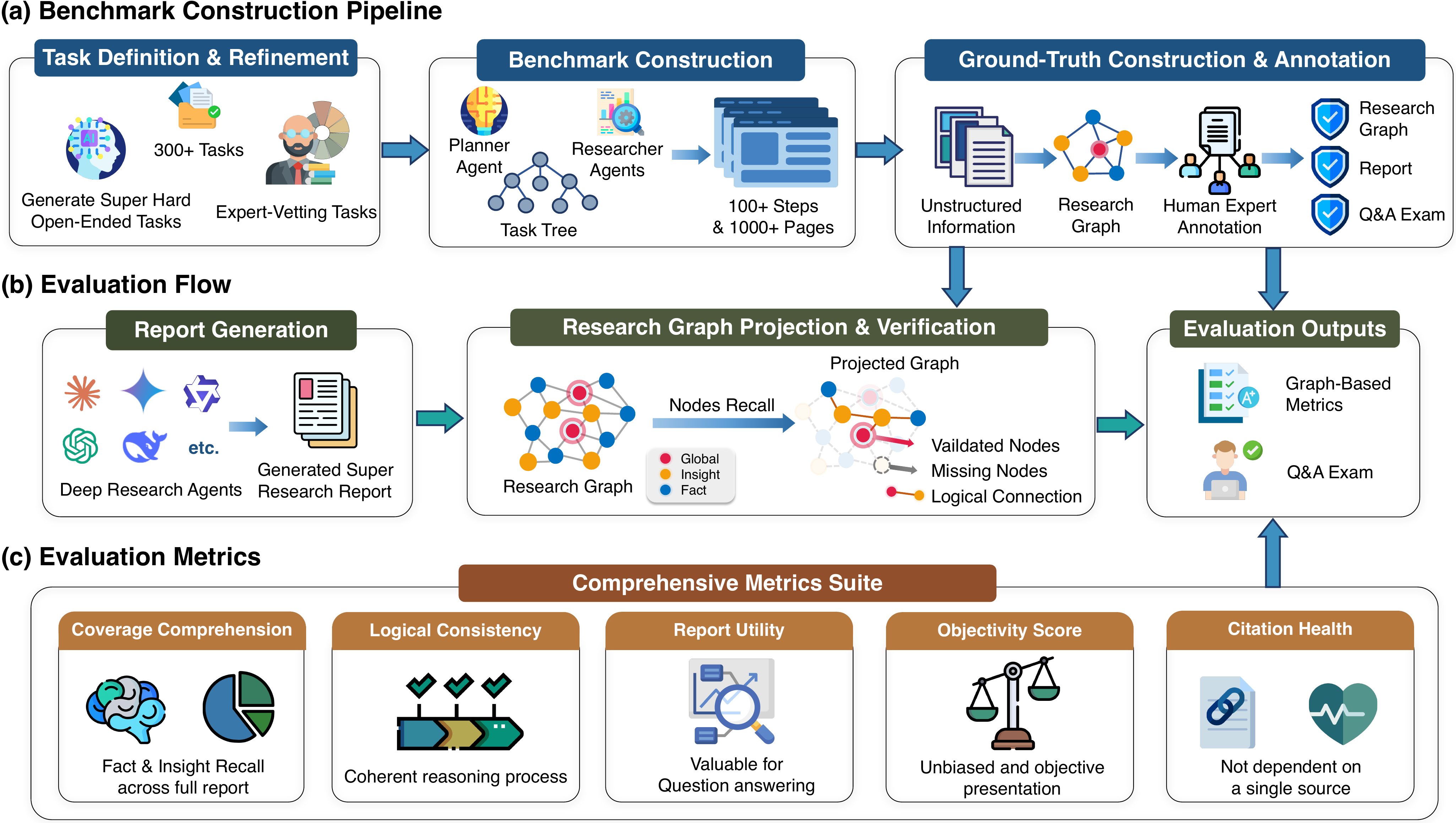}
    \caption{
        Overview of the SuperResearch Benchmark framework. \textbf{(a) Construction Pipeline:} The process starts with the joint definition of 300+ ``super hard'' open-ended tasks, which undergo rigorous expert vetting. Autonomous agents then execute a long-horizon research process involving 100+ retrieval steps and the synthesis of 1,000+ web pages. The resulting ``Gold Standard'' consists of a structured Research Graph, canonical reports, and a question-answer (QA) exam. \textbf{(b) Evaluation Flow:} Research reports are audited via Research Graph Projection. The system maps claims from the generated report onto the ground-truth Research Graph to verify Nodes Recall (categorized into atomic facts and insights) and the integrity of Logical Connections, ensuring high-level conclusions are grounded in verifiable evidence. \textbf{(c) Metrics Suite:} A comprehensive five-dimensional suite quantifies model performance, including Coverage \& Comprehension ($\mathcal{R}_\text{weighted}$), Logical Consistency ($\mathcal{C}_\text{logic}$), Report Utility ($\mathcal{U}_\text{qa}$), Objectivity Score ($\mathcal{O}_\text{bias}$) and Citation Health.  
    }
    \label{fig:metric_validation}
\end{figure*}

Although highly complex, multi-faceted queries may be infrequent in routine consumer applications, Super Research serves a vital role as a frontier stress test and a high-ceiling evaluation framework for LLMs. 
As current benchmarks begin to suffer from performance saturation, there is a growing need for ``ceiling protocols'' that push models to their absolute operational limits. 
By demanding long-horizon planning and the orchestration of over 100+ retrieval steps and 1,000+ web pages, Super Research exposes latent weaknesses in reasoning consistency and context management that simpler tasks fail to trigger. Consequently, a model's proficiency within this framework functions as a powerful proxy for its general research competence; achieving success in such a high-entropy environment suggests a level of algorithmic robustness and ``agentic''  stability necessary to navigate virtually any subordinate or specialized research task with high reliability. 
\begin{figure*}[t]
    \centering
    \includegraphics[width=0.99\linewidth]{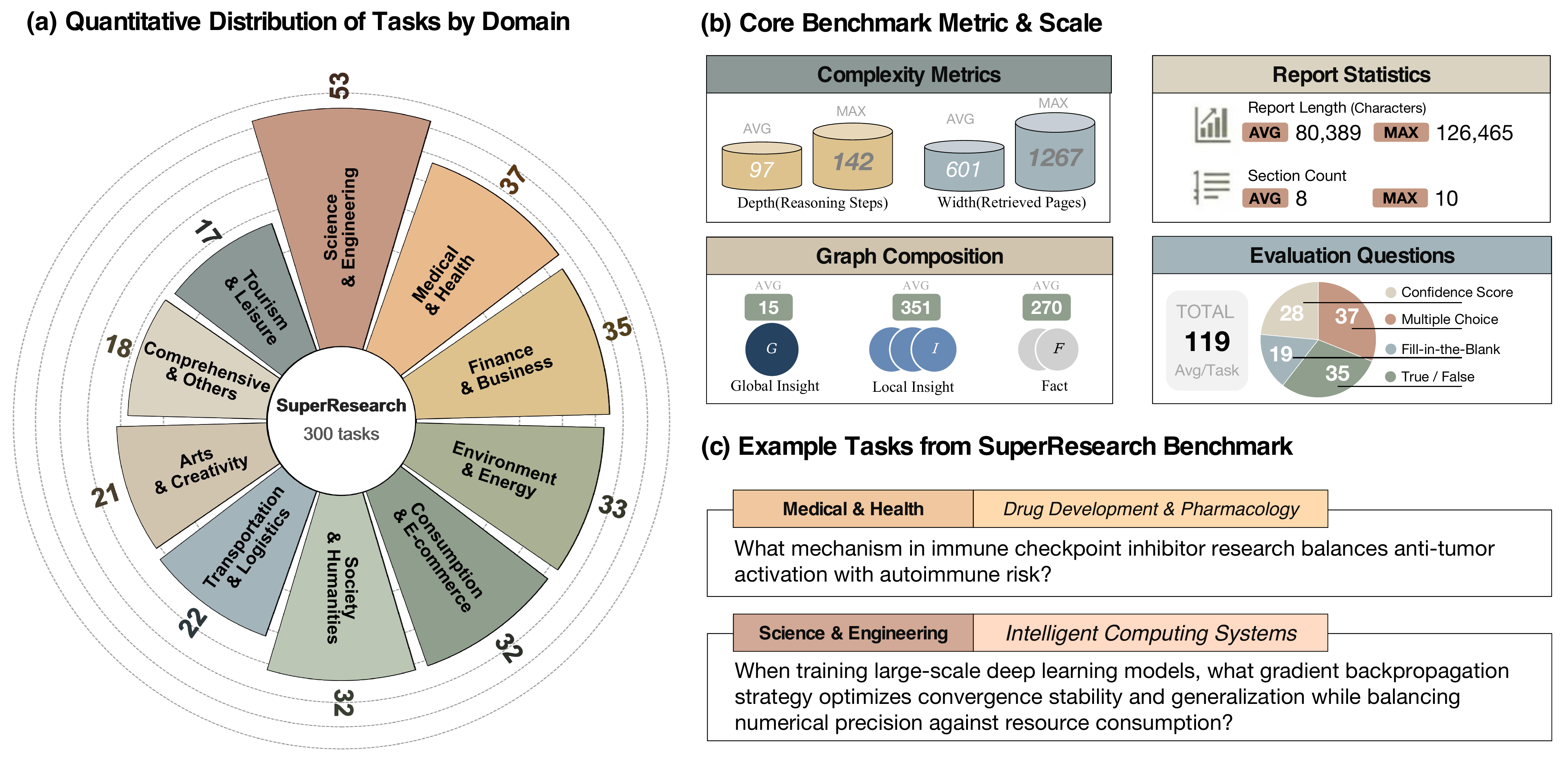}
    \caption{
        Structural and Functional Characterization of the SuperResearch Benchmark.
        \textbf{(a) Quantitative Distribution of Tasks by Domain:} A Rose Chart illustrating the distribution of 300 expert-written tasks across 10 specialized domains.
        \textbf{(b) Core Benchmark Metrics \& Scale:} Quantitative statistics characterizing the ``ceiling-level'' challenge across four key dimensions: 
        \textit{Complexity Metrics} (measuring reasoning depth and retrieval breadth), 
        \textit{Report Statistics} (tracking content volume and structure), 
        \textit{Graph Composition} (quantifying hierarchical knowledge density), and 
        \textit{Evaluation Questions} (showing the diversity of audit mechanisms).
        \textbf{(c) Example Tasks from SuperResearch Benchmark:} Representative inquiries exhibiting multi-objective trade-offs and conflicting evidence, serving as a ``ceiling-level'' challenge.
    }
    \label{fig:data_dist}
\end{figure*}

\section{The SuperResearch Benchmark}\label{sec:data}
In this section, we detail the methodology used to construct the \textsc{SuperResearch} benchmark. We employ an expert-involved, axiomatic construction process aiming to establish a ``Gold Standard'' for deep research tasks characterized by high cognitive complexity and rigorous logical structures. The core design philosophy of \textsc{SuperResearch} lies in achieving a dual balance: the breadth of information coverage (\textit{SuperWide}) and the depth of synthetic reasoning (\textit{SuperDeep}). Unlike existing datasets that focus on fact retrieval, our benchmark is engineered to test the asymptotic limits of an agent's ability to navigate unstructured and high-entropy information landscapes.

\subsection{Task Definition}\label{sec:data-task}
To stress-test the upper limits of Deep Research Agents, we synthesized criteria from frontier models and domain experts to establish a rigorous design protocol. Our construction follows three core principles: \ding{172} \textit{Frontier-Level Specificity} targeting unresolved challenges intelligible to domain experts and excluding general-knowledge queries to ensure the benchmark measures deep expertise rather than surface-level reasoning. \ding{173} \textit{Search-Intensive Complexity} demands broad, multi-hop synthesis across heterogeneous sources. Beyond simple fact retrieval, tasks require the analysis of conflicting data and long causal chains to test information integration. \ding{174} \textit{Professional Precision} employs unambiguous, industry-standard terminology to define the research scope, ensuring performance gaps reflect genuine capability limitations rather than interpretation errors.

\subsection{Expert Refinement}\label{sec:data-principle}
We employ a top-down, expert-driven construction pipeline. As illustrated in Figure \ref{fig:metric_validation}, our process integrates the generative breadth of LLMs with the discerning judgment of human domain experts to ensure high cognitive density. The construction process consists of two distinct phases:

To generate high-complexity programs, we utilized advanced LLMs (GPT-4o, Gemini-deep-research-agent) under our \textit{Cognitive-Rank Constrained Expert Simulation} framework. By injecting strict negative constraints to enforce a senior investigator persona, models were instructed to: \ding{172} traverse sub-domain topologies to identify structural tensions; \ding{173} prioritize multi-layered analysis over factual lookups; and \ding{174} scale task volume by theoretical field density. 
Subsequently, domain experts (senior practitioners and Ph.D. candidates) audited the raw candidates to eliminate hallucinations. The screening retained only tasks meeting three criteria: \ding{172} theoretical solvability via public retrieval; \ding{173} genuine complexity resistant to simple keyword search; and \ding{174} unambiguous formulation, ensuring alignment with real-world research demands.




\subsection{Human AI collaborative Collection}
\label{sec:data-collection}

Our four-stage dataset construction pipeline is designed to mimic the cognitive workflow of a domain expert. This process integrates autonomous agent collaboration with human-in-the-loop verification to ensure both the breadth of information retrieval and the logical depth of synthesis.

\textit{Phase 1: Hierarchical Task Decomposition and Retrieval.}
This phase addresses the challenge of information breadth and logical dependency. Given a complex research query, a \textit{Planner} agent \ref{app:planner} decomposes the root topic into a structured, Directed Acyclic Graph (DAG) of research tasks, organized into chronological phases and specific investigation chapters.
To achieve progressive generation, the \textit{Researcher} \ref{app:researcher} and \textit{Summarizer} \ref{app:summarizer} operate in a collaborative loop: as the Researcher executes sub-tasks in a dependency-aware sequence, the Summarizer continuously synthesizes the results into a \textit{Dynamic Memory}. This memory is iteratively injected into subsequent steps, ensuring that the global context evolves alongside the execution.

\textit{Phase 2: Research Graph Construction.}
We transform unstructured sub-reports into a structured research graph through a collaborative Human-AI process, as shown in Figure~\ref{fig:graph_construction}. This phase integrates subgraph extraction, expert-guided merging, and global insight generation. First, the system extracts atomic ``Facts'' (anchored to specific URLs) and derived ``Insights'' from sub-reports. Subsequently, adopting a human-in-the-loop workflow, experts refine the graph structure, constructing a bottom-up topology where higher-order insights are rigorously connected to supporting factual nodes. This verification is augmented by LLM-driven code sandboxes and search tools for automated calculation and fact-checking~\ref{app:graph_gen}. Finally, both experts and the LLM traverse this graph to identify evidence clusters, iteratively generating global insights that bridge disparate sub-reports.

\textit{Phase 3: Report Synthesis and Expert Editing.}
This phase focuses on producing definitive canonical reports. Leveraging the structured research graph and granular sub-reports, a \textit{Writer} agent \ref{app:writer} iteratively constructs the manuscript section by section. Crucially, this process integrates a real-time human-in-the-loop workflow: domain experts interact via a dashboard to validate logical soundness and intervene with on-the-fly corrections during text generation.

\textit{Phase 4: Automated Evaluation Metric Construction.}
Leveraging both the research graph and the final report, we derive a comprehensive suite of Question-Answer (QA) pairs \ref{app:metric}. Beyond standard fact retrieval, this suite specifically includes ``Bias Calibration'' queries designed to assess the neutrality and objectivity of the report's narrative, ensuring the description remains impartial even when synthesizing subjective or conflicting sources. 
The generated QA pairs and calibrated bias scores serve as the ground truth for benchmarking the \textit{SuperDeep} and \textit{SuperWide} capabilities of the models evaluated in this study.

\section{Evaluation Framework and Metrics}\label{sec:eval}

Evaluating frontier-level, open-ended research poses unique challenges. Unlike factoid retrieval, open-ended queries allow models to explore divergent research trajectories, rendering traditional string-matching metrics ineffective. Furthermore, frontier questions often lack static ground truths, necessitating a deeper assessment of the reasoning process and derivation depth rather than mere surface-level overlap.

To address these challenges, we introduce a comprehensive evaluation suite anchored in rigorously constructed Research Graphs. 
Specifically, this framework assesses report quality across five complementary dimensions: \ding{172} Coverage and Comprehension (graph-based information recall), \ding{173} Logical Consistency (coherence of reasoning chains), \ding{174} Report Utility (QA-based functional validation), \ding{175} Objectivity Score (viewpoint neutrality assessment), and \ding{176} Citation Health (source diversity and dependency analysis). Finally, the Overall Score is calculated as a weighted average of the first four dimensions, with Citation Health serving as a separate diagnostic metric.



\subsection{Coverage and Comprehension}

To comprehensively assess both the breadth and depth of the generated reports, we introduce a metric based on research graph projection (as shown in Figure~\ref{fig:intro}). The core principle involves projecting the ground-truth graph structure onto the generated text to verify the presence of essential information. We first calculate the recall rate for individual graph nodes, categorizing them into three hierarchical levels: Atomic Facts Level ($\mathcal{R}_{\text{1}}$) for specific data points, Key Insights Level ($\mathcal{R}_{\text{2}}$) for intermediate reasoning connections, and Global Insights Level ($\mathcal{R}_{\text{3}}$) for high-level strategic conclusions. To derive a single, holistic indicator, we compute the Depth-Weighted Recall (denoted as $\mathcal{R}_{\text{weighted}}$). This metric assigns progressive weights to nodes based on their hierarchical depth, ensuring that the evaluation prioritizes the capture of complex synthesis and core theses over the mere enumeration of trivial facts. The detailed mathematical formulation is provided in Appendix~\ref{app:kc_formulation}.

\subsection{Logical Consistency}
Research fundamentally demands that conclusions be derived from evidence, not merely stated. To verify the integrity of these deduction paths, we propose Logical Consistency (denoted as $\mathcal{C}_{\text{logic}}$).  This metric assesses whether the generated Global Conclusions ``are algorithmically grounded in Atomic Facts'' via complete, unbroken citation chains. Formally, $\mathcal{C}_{\text{logic}}$ is defined as the product of the valid inference ratio (the proportion of conclusions fully supported by valid evidence edges) and the overall graph coverage. This multiplicative formulation serves as a rigorous filter: it penalizes reports that hallucinate correct high-level answers without the necessary supporting evidence, ensuring that high scores reflect both accurate outcomes and valid reasoning processes. The mathematical derivation is detailed in Appendix~\ref{app:lc_formulation}.

\subsection{Report Utility}
Beyond structural completeness, the ultimate measure of a research report is its tangible value to domain experts. We introduce Report Utility (denoted as $\mathcal{U}_{\text{qa}}$), which evaluates whether the generated content successfully encodes actionable knowledge. We construct a dynamic set of exam questions derived directly from the \textit{Reports}. The evaluation is conducted as a \textbf{closed-context reading comprehension task}: an independent LLM judge attempts to answer these questions relying \textit{exclusively} on the information explicitly present in the generated report. This rigorous constraint ensures that high scores are awarded only when the report serves as a self-contained, high-fidelity information source, effectively penalizing hallucinations or vague generalizations. The detailed evaluation protocol is provided in Appendix~\ref{app:qa_formulation}.

\subsection{Objectivity Score}
For complex, open-ended research topics, a high-quality report must navigate conflicting viewpoints without succumbing to one-sided bias. To measure this, we utilize the \textbf{Objectivity Score} (denoted as $\mathcal{O}_{\text{bias}}$), which quantifies the model's ability to maintain \textbf{multi-perspective balance}.

Instead of verifying binary truth, this metric conducts a double-blind audit to assess how the report weighs competing arguments (e.g., Thesis vs. Antithesis). The core objective is \textbf{Calibration of Stance}: the evaluation ensures that the degree of certainty and the distribution of evidence in the generated report accurately mirror the inherent ambiguity found in the ground-truth Research Graph. A high $\mathcal{O}_{\text{bias}}$ indicates that the model successfully represents the diversity of perspectives, avoiding the pitfall of presenting a highly debated topic as a settled fact. The detailed calculation methodology is provided in Appendix~\ref{app:os_formulation}.


\subsection{Citation Health}
While not part of the primary score, Citation Health serves as a diagnostic measure of \textbf{information synthesis quality}. Its primary objective is to distinguish between genuine multi-source integration and mere single-document summarization, flagging potential risks of Single-Source Dependency. We monitor two diagnostic indicators (where lower is better): \ding{172} \textit{Source Dominance} ($\mathcal{D}_{\text{src}}$) checks the total volume, measuring if a single reference constitutes a disproportionately high percentage of the total citations. 
\ding{173} \textit{Narrative Monopolization} ($\mathcal{M}_{\text{mono}}$) evaluates the citation layout across the full text. Unlike localized clusters (where using one source for a specific paragraph is acceptable), this metric penalizes cases where a single source drives the narrative flow from beginning to end, implying the report is merely a summarization of one document rather than a multi-source synthesis. Formal definitions are in Appendix~\ref{app:reliability_formulation}.

\section{Experiments}\label{sec:exper}
\subsection{Experimental Setup}\label{sec:exper-setup}
\begin{table*}[t]
    \centering
    \scriptsize
    \caption{Benchmarking the Performance Landscape of Research Agents on SuperResearch Benchmark. 
        Evaluation of 12 representative research systems across three architectural paradigms: 
        \colorbox{blue!8}{Deep Research System}, 
        \colorbox{orange!8}{Native Search-Integrated Agent}, and \colorbox{green!8}{Search-Augmented Baseline}.
        \best{Bold} and \secBest{underline} denote the best and second-best performance, respectively; $\uparrow$ ($\downarrow$) indicates that higher (lower) values are preferred.
        \metric{Overall}: Aggregate score synthesized from the four core metrics (\metric{Coverage}, \metric{Consistency}, \metric{Utility}, and \metric{Objectivity}).
        \metric{Coverage} ($\mathcal{R}_{\text{weighted}}$): Coverage \& Comprehension; 
        \metric{Consistency} ($\mathcal{C}_{\text{logic}}$): Logical Consistency; 
        \metric{Utility} ($\mathcal{U}_{\text{qa}}$): Report Utility; 
        \metric{Objectivity} ($\mathcal{O}_{\text{bias}}$): Objectivity Score.
        \metric{Citation Health}: Comprises \metric{Dominance} ($\mathcal{D}_{\text{src}}$, Source Dominance) and \metric{Monopolization} ($\mathcal{M}_{\text{mono}}$, Narrative Monopolization).
    }
    \vspace{-0.8em} 
    \renewcommand{\arraystretch}{1.2} 
    \resizebox{\textwidth}{!}{
    \begin{tabular}{l ccccc cc}
        \hline
        \rowcolor{gray!6} & & & & & & \multicolumn{2}{c}{\textbf{Citation Health}} \\
        \cline{7-8}
        \rowcolor{gray!6} \multirow{-2}{*}{\textbf{Method}}
        & \multirow{-2}{*}{\makecell{\textbf{Overall} \\ }~$\uparrow$}
        & \multirow{-2}{*}{\makecell{\textbf{Coverage} \\ {\scriptsize ($\mathcal{R}_{\text{weighted}}$)}}~$\uparrow$}
        & \multirow{-2}{*}{\makecell{\textbf{Consistency} \\ {\scriptsize ($\mathcal{C}_{\text{logic}}$)}}~$\uparrow$}
        & \multirow{-2}{*}{\makecell{\textbf{Utility} \\ {\scriptsize ($\mathcal{U}_{\text{qa}}$)}}~$\uparrow$}
        & \multirow{-2}{*}{\makecell{\textbf{Objectivity} \\ {\scriptsize ($\mathcal{O}_{\text{bias}}$)}}~$\uparrow$}
        & \makecell{\textbf{Dominance} \\ {\scriptsize ($\mathcal{D}_{\text{src}}$)}}~$\downarrow$ & \makecell{\textbf{Monopolization} \\ {\scriptsize ($\mathcal{M}_{\text{mono}}$)}}~$\downarrow$ \\
        \hline

        \rowcolor{blue!8} \multicolumn{8}{c}{Deep Research System} \\
        \hline
        Gemini Deep Research & \best{28.62} & \best{33.15} & \secBest{22.92} & \secBest{22.98} & \secBest{35.43} & 4.85 & 39.71 \\
        Sonar Deep Research & \secBest{27.04} & 28.26 & 16.89 & \best{26.50} & \best{36.50} & 8.32 & 42.49 \\
        Tongyi Deep Research & 25.74 & 27.41 & 19.80 & 20.86 & 34.88 & - & - \\
        o3 Deep Research & 22.04 & 23.44 & 19.90 & 13.66 & 31.18 & 5.68 & \best{32.33} \\
        o4-mini Deep Research & 21.91 & 23.11 & 20.04 & 13.60 & 30.91 & 7.35 & \secBest{36.50} \\
        \hline

        \rowcolor{orange!8} \multicolumn{8}{c}{Native Search-Integrated Agent} \\
        \hline
        Kimi-k2 & 26.16 & 25.97 & \best{24.80} & 19.88 & 34.01 & - & - \\
        Grok-4-1-fast & 24.55 & \secBest{30.09} & 19.41 & 16.79 & 31.91 & 9.33 & 43.14 \\
        \hline
        
        \rowcolor{green!8} \multicolumn{8}{c}{Search-Augmented Baseline} \\
        \hline
        Deepseek-r1 & 22.57 & 24.21 & 15.81 & 16.55 & 33.73 & \secBest{2.94} & 55.09 \\
        Claude-3.5-sonnet & 22.26 & 27.02 & 17.60 & 13.75 & 30.69 & \best{2.08} & 65.42 \\
        Minimax-m2.1 & 22.05 & 20.91 & 13.24 & 19.24 & 34.81 & 10.54 & 44.56 \\
        Llama-3.3-70B & 17.71 & 15.55 & 13.61 & 11.44 & 30.25 & 57.51 & 62.66 \\
        Qwen-2.5-72B & 16.84 & 10.47 & 10.78 & 14.72 & 31.39 & - & - \\
        \hline
    \end{tabular}
    }
    \label{tab:main_results}
\end{table*}


\noindent
\textbf{Evaluation Methods.}
To comprehensively benchmark the capabilities of current systems in super research tasks, we evaluate a diverse set of methods. The evaluated methods are grouped into three primary tiers:

\noindent
(1) \textit{Deep Research Systems}. 
 We assess leading commercial agents explicitly optimized for autonomous, long-horizon research tasks. These systems are designed to handle complex planning, multi-step evidence gathering, and information synthesis natively. Our evaluation includes Sonar Deep Research~\cite{perplexity2025sonar}, Gemini Deep Research~\cite{google2025deepresearch}, Tongyi Deep Research~\cite{qwen2024technical} and OpenAI's latest reasoning-enhanced models, o3 Deep Research and o4-mini Deep Research~\cite{openai2025deepresearch}. 

\noindent
(2) \textit{Native Search-Integrated Agents}.
To examine the capabilities of models with integrated real-time web access but without the full ``Deep Research'' agentic loop, we evaluate Kimi-k2~\cite{moonshot2025kimik2} and Grok-4-1-fast~\cite{xai2025grok4}. These models leverage their native search tools to answer complex queries directly, serving as a strong baseline for integrated search-augmented generation.

\noindent
(3) \textit{Search-Augmented Baselines}. 
To isolate the reasoning capabilities of foundation models from search infrastructures, we construct a standardized agentic workflow using the LangGraph framework~\cite{langgraph} equipped with Tavily Search~\cite{tavily} as the unified retrieval tool. We integrate a diverse set of high-performing LLMs, including DeepSeek-r1~\cite{deepseek2025r1}, MiniMax-m2.1~\cite{minimax2025m21}, Claude-3.5-sonnet~\cite{anthropic2024claude35}, Qwen-2.5-72B~\cite{qwen2024technical}, and Llama-3.3-70B~\cite{meta2024llama33}. All evaluations are conducted under controlled conditions to ensure fair comparison across different architectures and deployment environments.

\subsection{Main Results}
\label{sec:exper-results}
\subsubsection{Overall Performance: The Challenge of Super-Complexity}
\textit{The results in Table~\ref{tab:main_results} indicate that the SuperResearch Benchmark poses a rigorous challenge to all current agent paradigms.} Even the state-of-the-art (SOTA) system, Gemini Deep Research, achieves an Overall Score of only 28.62. This relatively low absolute ceiling confirms that super-complex queries 
remain a largely unsolved frontier.

The \textbf{Deep Research System} tier generally leads, with Gemini Deep Research achieving the highest average (28.62), followed by Sonar Deep Research (27.04), securing the top positions.
However, Tongyi Deep Research (25.74) and OpenAI's o3/o4-mini Deep Research ($\sim$22) show a clear divergence from the leaders.
Surprisingly, the \textbf{Native Search-Integrated Agent} Moonshot Kimi-k2 (26.16) and Grok (24.55) outperform the search-augmented baseline models and surpass part of the specialized Deep Research systems.
In contrast, the \textbf{Standard Baseline} tier, including Minimax-m2.1 and DeepSeek-r1, clusters in the 16--23 score range. Despite utilizing powerful foundation models, these framework-based approaches lag behind the top commercial systems, highlighting the persistent gap in system integration for complex research tasks.

\subsubsection{Coverage and Consistency}
\textit{Our data indicates a positive correlation between retrieval breadth and reasoning depth, suggesting that extensive Coverage is a vital foundation for Consistency.}

\ding{172} \textbf{The Reasoning Bottleneck \& Retrieval Dependency.} Logical Consistency ($\mathcal{C}_{\text{logic}}$) appears to represent a universal performance bottleneck, current systems \textit{may be predominantly constrained to surface-level information aggregation}, often failing to progress toward deeper, multi-hop logical synthesis (Table~\ref{app:coverage}). Furthermore, we observe a potential dependency where limited retrieval coverage (e.g., $\mathcal{R}_{\text{weighted}} < 16$ for LangGraph models) seems to impose a ceiling on reasoning performance ($\mathcal{C}_{\text{logic}} < 14$). However, even systems with superior coverage (such as Gemini, $\mathcal{R}_{\text{weighted}} \approx 33.15$) do not proportionally breakthrough this reasoning barrier ($\mathcal{C}_{\text{logic}} \approx 22.92$), indicating that both high-quality retrieval and analysis need further optimization (see Appendix~\ref{app:case_failure_lack_consistency}).

\ding{173} \textbf{Variance in Logical Alignment.} Research reports exhibit distinct structural disparities in bridging evidence to conclusions. Kimi-k2 achieves Logical Consistency ($\mathcal{C}_{\text{logic}} \approx 24.80$) relative to its Coverage ($\mathcal{R}_{\text{weighted}} \approx 25.97$). In contrast, Grok-4-1-fast captures broader information ($\mathcal{R}_{\text{weighted}} \approx 30.09$) but struggles to organize this data into a logically sound narrative ($\mathcal{C}_{\text{logic}} \approx 19.41$). This dissociation highlights that high recall alone does not ensure the structural integrity required to align facts with rigorous insights.

\subsubsection{The Utility-Objectivity Trade-off}
\textit{The Report Utility ($\mathcal{U}_{\text{qa}}$) and Objectivity ($\mathcal{O}_{\text{bias}}$) metrics reveal significant divergences in model alignment strategies.}

\ding{172} \textbf{Defensive Summarization Paradox.} OpenAI's o3/o4-mini Deep Research exhibit a distinct ``defensive posture'', combining high Objectivity ($\mathcal{O}_{\text{bias}} > 30$) with the lowest Utility scores ($\mathcal{U}_{\text{qa}} \approx 14$). This pattern suggests a tendency to prioritize safe, generalized summaries, potentially sacrificing the granular details required for expert-level queries (see Appendix~\ref{app:case_conservatism}).

\ding{173} \textbf{The Objectivity Ceiling.} Notably, most models converge within a narrow high-score range ($\mathcal{O}_{\text{bias}} > 30$). While this reflects effective baseline safety, the lack of breakthrough scores suggests that current agents still struggle to navigate complex, conflicting evidence with expert-level nuance.

\subsubsection{Citation Health Diagnostics}
\textit{Reliability metrics reveal distinct structural patterns in information synthesis beyond simple citation counts. Lower scores for both Source Dominance ($\mathcal{D}_{\text{src}}$) and Narrative Monopolization ($\mathcal{M}_{\text{mono}}$) indicate more balanced research.}

\ding{172} \textbf{Structural Over-Reliance.} Models in the LangGraph tier exhibit structural weaknesses driven by low citation volume. Llama-3.3-70B shows poor performance across both metrics ($\mathcal{D}_{\text{src}}  \approx  57.51, \mathcal{M}_{\text{mono}}  \approx  62.66$), primarily because it retrieves extremely few unique sources and repeatedly cites them, failing to broaden its evidence base (see Appendix~\ref{app:case_overreliance}). 

\ding{173} \textbf{Sparse Citation Recycling.} Similarly, while Claude-3.5-sonnet and DeepSeek-r1 achieve low Dominance scores ($\mathcal{D}_{\text{src}} < 3$), their seemingly even distribution is based on a sparse number of total citations. These models frequently recycle a small pool of references throughout the text to sustain the narrative, resulting in high Monopolization scores ($\mathcal{M}_{\text{mono}} > 55$) that reflect a lack of informational diversity (see Appendix~\ref{app:case_overload}).

\subsection{Ablation Study}\label{sec:exper-ablation}
\subsubsection{Sensitivity Analysis: Graph-Based Metrics vs. LLM-as-a-Judge}
To validate the reliability, we conducted a comparative study against the standard ``LLM Judge'' baseline using 120 reports randomly generated from different tasks and models. We designed a controlled perturbation experiment to measure evaluator responsiveness in two scenarios: \ding{172} \textit{Degradation:} Removing 1$\sim$3 key facts or insights from the original report; \ding{173} \textit{Improvement:} Injecting 1$\sim$3 relevant facts into appropriate sections (Details in~\ref{app:ablation_prompts}). 
We computed the \textit{Responsiveness Rate} (formal definition in Appendix~\ref{app:responsiveness}), defined as the percentage of instances where the metric's score correctly shifts in alignment with the perturbation (i.e., decreases for degradation and increases for improvement).

\begin{figure}[t]
    \centering
    \begin{minipage}[b]{0.48\textwidth}
        \centering
        \includegraphics[width=\linewidth]{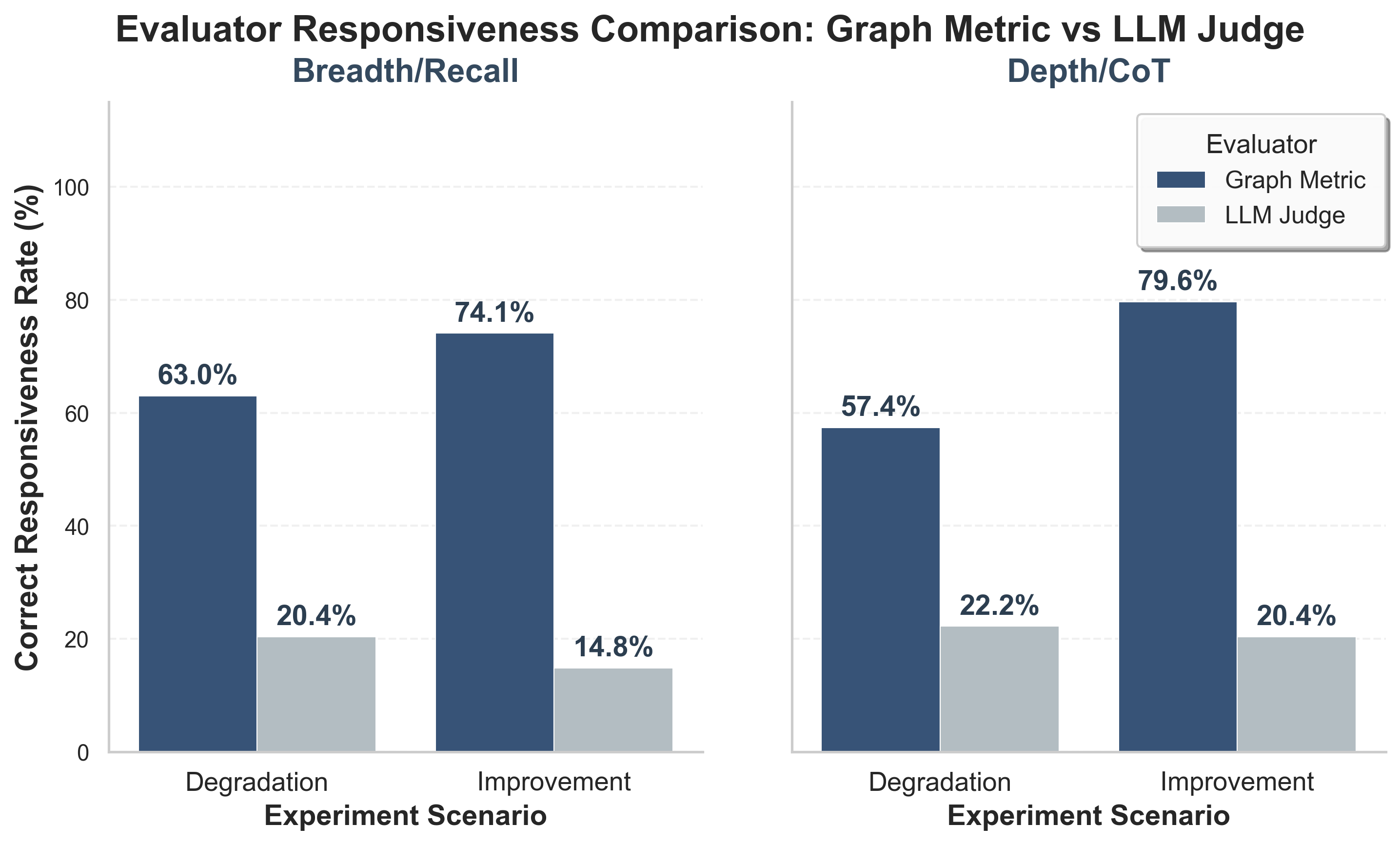}
        \caption{Evaluation Sensitivity Analysis. Compared to the LLM Judge (grey), our Graph Metric (dark blue) shows superior responsiveness to quality fluctuations.}
        \label{fig:responsiveness}
    \end{minipage}
    \hfill 
    \begin{minipage}[b]{0.48\textwidth}
        \centering
        \scriptsize
        \renewcommand{\arraystretch}{1.1} 
        \setlength{\tabcolsep}{6pt} 
        \begin{tabular}{llc}
            \toprule
            \textbf{Metric} & \textbf{Evaluator} & \textbf{Norm STD} $\downarrow$ \\
            \midrule
            \multirow{2}{*}{Coverage} & Ours Graph & 4.42\% \\
                                    & LLM Judge & \textbf{3.88\%} \\
            \midrule
            \multirow{2}{*}{Consistency} & Ours Graph & \textbf{0.83\%} \\
                                       & LLM Judge & 4.69\% \\
            \midrule
            \multirow{2}{*}{Utility} & Ours Q\&A & \textbf{2.05\%} \\
                                   & LLM Judge  & 4.37\% \\
            \midrule
            \multirow{2}{*}{Overall} & Ours Method & \textbf{1.67\%} \\
                                   & LLM Judge  & 3.02\% \\
            \bottomrule
        \end{tabular}
        \captionof{table}{Evaluator Consistency Analysis. $\sigma_{norm}$ indicates the fluctuation as a percentage (lower is better).}
        \label{tab:consistency}
    \end{minipage}
\end{figure}
The experimental results, as illustrated in Figure \ref{fig:responsiveness}, highlight a significant and observable performance gap between the methods. The \textbf{Graph Metric} demonstrated exceptional sensitivity to qualitative changes, consistently achieving responsiveness rates ranging from 57.4\% to 79.6\% across both Breadth and Depth dimensions. In contrast, the \textbf{LLM Judge} struggled to distinguish these fine-grained perturbations, as evidenced by its responsiveness rates fluctuating widely between 14.8\% and 22.2\%. This demonstrates the robustness of the graph metric.

\subsubsection{Evaluator Consistency: Robustness to Model Bias}
To assess objectivity, we analyzed the variance in scores across different LLM evaluators. We evaluated a random sample of 120 reports using three distinct models (Gemini-2.5-Flash, GPT-4o-mini, and Qwen2.5-72B-Instruct) to serve as both the LLM (Details in~\ref{prompt:comprehensive_eval}) and the parser for our Graph Metric. We posit that a robust evaluation metric should remain consistent, regardless of the underlying model. We employ range-normalized standard deviation ($\sigma_{norm}$) to compare consistency across metrics with different scales (e.g., 0-100 for Graph metrics vs. 1-10 for LLM judges). As shown in Table \ref{tab:consistency}, our Graph-based method demonstrates greater consistency across most metrics.

The results indicate a fundamental advantage in determinism for the Graph-based approach. For example, in the \textit{Objectivity} dimension, the Graph metric exhibits a negligible fluctuation of \textbf{1.07\%}, while the corresponding LLM Judge metric fluctuates by \textbf{7.82\%}—a nearly $65\times$ increase in variance. Similarly, for \textit{Depth} and \textit{Utility}, the Graph metrics reduce relative variance by approximately $6\times$ and $2.5\times$. Although different models may vary slightly in their information extraction capabilities, grounding of scores in a verifiable Research Graph and QA exams ensures that the final evaluation remains stable and reproducible.
\section{Conclusion}



We introduce \textbf{SuperResearch Bench}, a ``ceiling-level'' paradigm integrating \textit{Super Wide Retrieval} and \textit{Super Deep Investigation}. Our evaluation of 300 expert tasks—each requiring up to 100+ retrieval steps and 1,000+ web pages to reconcile conflicting evidence—via Graph-Anchored Auditing reveals that SOTA systems score under 29\%, failing to reconcile extreme breadth with logical depth. As standard benchmarks saturate, Super Research serves as a vital stress test to drive the evolution of robust, long-horizon autonomous deep research agents.

\section*{Impact Statements}
This work advances the field of autonomous research by establishing a rigorous testbed for evaluating high-complexity intelligence. By defining standards for ``strategic-tier'' synthesis across professional intelligence, scientific discovery, and strategic planning, we provide a necessary guide for future developments in agentic AI. However, as agents strive to synthesize massive evidence to meet these standards, there is a risk of compounding hallucinations, which may induce dangerous human over-reliance on flawed but authoritative-sounding reports. Furthermore, the high computational cost of such long-horizon retrieval necessitates a parallel focus on efficiency and ``Green AI'' practices.

\bibliographystyle{nips}  
\small
\bibliography{nips}
\normalsize


\newpage
\appendix
\onecolumn
\appendixpage

\begin{spacing}{1}
	\section*{Contents}
	\startcontents[appendices]
	\printcontents[appendices]{}{-1}{\setcounter{tocdepth}{2}}
\end{spacing}

\clearpage
\section{Related Work}

\paragraph{Autonomous Research Agent Frameworks.}
Autonomous research agents have evolved from basic reasoning techniques to sophisticated multi-agent systems. Foundational approaches such as Chain-of-Thought~\cite{wei2022chain} and ReAct~\cite{yao2022react} established core mechanisms for task decomposition and tool integration, operationalized in early systems like WebGPT~\cite{nakano2021webgpt} and WebCPM~\cite{qin2023webcpm}. Framework libraries, including LangChain~\cite{langchain} and LangGraph~\cite{langgraph} formalized complex, stateful reasoning workflows. Multi-agent paradigms, exemplified by AutoGen~\cite{wu2024autogen} and MetaGPT~\cite{hong2023metagpt}, introduced collaborative role-based architectures for intricate research processes. Specialized systems like STORM~\cite{shao2024assisting} for open-domain synthesis and ChemCrow~\cite{bran2023chemcrow} for scientific discovery advanced long-form research automation. However, a fundamental limitation persists: existing agents typically optimize for either breadth of information gathering or depth of verification (via mechanisms like Reflexion~\cite{shinn2023reflexion}), but rarely achieve both simultaneously. Recent efforts such as DeepWideSearch~\cite{deepwidesearch} and Step-DeepResearch~\cite{hu2025step} address this trade-off but remain constrained by limited planning horizons and insufficient robustness for high-entropy, conflicting information landscapes.

\paragraph{Knowledge Synthesis and Retrieval Methods.}
Information retrieval has evolved from simple vector-based similarity to structured knowledge integration paradigms. Early Retrieval-Augmented Generation (RAG) approaches~\cite{lewis2020retrieval,guu2020retrieval} mitigated hallucination but struggled with contextual fragmentation across disparate sources. Recent graph-based methods, including GraphRAG~\cite{edge2024local}, HiPPoRAG~\cite{jimenez2024hipporag}, and LightRAG~\cite{jiang2024lightrag}, leverage structured representations for multi-hop reasoning and heterogeneous information aggregation. By incorporating schema-based grounding~\cite{pan2024unifying,wu2023graph}, these approaches model semantic relationships between sources, enabling synthesis of conflicting narratives. Unlike traditional RAG systems treating documents as isolated chunks, graph-anchored methods capture evidence topology, essential for resolving contradictory information streams.

\paragraph{Evaluation Frameworks and Quality Metrics.}
Assessing open-ended research outputs requires metrics beyond surface-level plausibility. General-purpose benchmarks like GAIA~\cite{gaia} and AgentBench~\cite{liu2023agentbench} provide broad evaluations but lack precision for detailed research quality assessment. Specialized frameworks, including DeepResearch Bench~\cite{deepresearchbench}, LiveResearchBench~\cite{wang2025liveresearchbench}, DeepResearch Arena~\cite{wan2025deepresearch}, and DeepResearchGym~\cite{coelho2025deepresearchgym}, address this gap. However, these rely predominantly on LLM-as-a-judge scoring~\cite{zheng2023judging} or atomic fact verification~\cite{min2023factscore}, failing to capture attributes like logical coherence, source diversity, and bias calibration. Super Research introduces a graph-anchored evaluation framework that systematically quantifies knowledge coverage, information density, and robustness for rigorous assessment of advanced reasoning capabilities.

\begin{table*}[t]
    \centering
    \caption{Comparison of Super Research with Existing Agentic Benchmarks. 
    Super Research represents a ``Ceiling-Level'' challenge compared to existing paradigms. While Wide Search focuses on horizontal data acquisition and DeepResearch Bench prioritizes vertical synthesis, Super Research integrates \textit{Super Wide Retrieval} and \textit{Super Deep Investigation}. Moreover, to address the limitations of shallow fact-recall metrics, we propose a \textbf{Graph-Anchored Auditing} protocol that can comprehensively evaluate reasoning depth, bias, and uncertainty. }
    \vspace{-0.5em}
    \label{tab:complexity_comparison}
    \renewcommand{\arraystretch}{1.35} 
    \setlength{\tabcolsep}{3pt} 
    
    \resizebox{\textwidth}{!}{
        \begin{tabular}{l l l l l l l}
            \toprule
            \multirow{2}{*}{\textbf{Dimensions}} & 
            \multirow{2}{*}{\textbf{Metric}} &
            \textbf{GAIA (Level 3)} & 
            \textbf{WideSearch} & 
            \textbf{DeepWideSearch} & 
            \textbf{DeepResearch Bench} & 
            \textbf{Super Research} \\
            
            & & \cite{gaia} & \cite{widesearch} & \cite{deepwidesearch} & \cite{deepresearchbench} & \textbf{(Ours)} \\
            \midrule
            
            \multirow{2}{*}{\textbf{Task Scope}} 
            & Core Paradigm & General Assistant & Broad Info-Seeking & Deep \& Wide Info-Seeking & Deep Research Agent & Super Research \\
            & Task Goal & Precise Problem Solving & Structured Aggregation & Complex Table Synthesis & Analyst-Grade Report Gen. & \textbf{Strategic Planning \& Discovery} \\
            \midrule

            \multirow{2}{*}{\textbf{Depth}} 
            & Steps / Iterations & 10--40 (Variable) & $\sim$1.2 / Entity & $\sim$4.2 / Entity & 10--20 (Iterative) & \textbf{$\sim$100 (Max 140+)} \\
            & Methodology & Tool Orchestration & Parallel Execution & Recursive / Hierarchical & Hierarchical Synthesis &Structured Decomposition \\
            \midrule

            \multirow{2}{*}{\textbf{Width}} 
            & Web Pages & $<$ 20 (Targeted Search) & $\sim$44 (High Width) & $\sim$414 Units (High Vol.) & $\sim$110 (Hundreds) & \textbf{$\sim$600 (Max 1,200+)} \\
            & Search Scope & Specific Factoid & Multi-Entity Coverage & Hidden Target Entities & PhD-Level Research Topic & Thematic Perspective Synthesis \\
            \midrule
            
            \multirow{2}{*}{\textbf{Evaluation}} 
            & Primary Method & Quasi-Exact Match & Hybrid (F1 \& SR) & Multi-Granularity F1 & Adaptive Ref-Based Judge & \textbf{Graph-Anchored Auditing} \\
            & Focus Dimensions & Result Correctness & Completeness \& Fidelity & Depth \& Width Integration & Quality \& Citation Accuracy & \textbf{Consistency, Bias \& Coverage} \\
            
            \bottomrule
        \end{tabular}
    }
\end{table*}

\clearpage
\section{More Details about SuperResearch Benchmark.}

\label{sec:more_details}

\subsection{Principles for Task Generation}
\label{app:principles}

\begin{reasoning_prompt}{Prompt: Cognitive-Rank Constrained Task Generation}
    \textbf{Role:} Senior Principal Investigator (NSF/ERC Level)\\
    \textbf{Objective:} Propose high-complexity research programs for the domain: \{domain\}.\\
    \textbf{Cognitive-Rank Constraints (Negative Constraints):}
    \begin{itemize}
        \item \textbf{NO "How-to" Guides:} Reject procedural or tutorial-style topics.
        \item \textbf{NO Fact Lookups:} Reject questions answerable by a single retrieval.
        \item \textbf{NO Settled Science:} Avoid topics with established consensus unless proposing a paradigm shift.
    \end{itemize}
    \textbf{Generation Instructions:}
    \begin{enumerate}
        \item \textbf{Traverse Topology:} Identify structural tensions and unresolved mechanisms.
        \item \textbf{Multi-Layered Analysis:} Proposals must require synthesizing economic, technical, and ethical layers.
        \item \textbf{Theoretical Density:} Generate more tasks for rapidly evolving fields.
    \end{enumerate}
\end{reasoning_prompt}

\subsection{System Prompts}
\label{sec:prompts}


\subsubsection{Planner Agent}
\label{app:planner}

\begin{reasoning_prompt}{System Prompt: Planner Agent (Hierarchical Decomposition)}
\textbf{Role:} Chief Research Architect\\
\textbf{Task:} According to the topic and expert advice. Create a MECE (Mutually Exclusive, Collectively Exhaustive) hierarchical research plan.\\
\textbf{Root Question:} \{root\_topic\}\\
\textbf{Structure Requirements:}
\begin{enumerate}
    \item \textbf{Phases:} Chronological or Logical progression (e.g., Background - Mechanism - Implication).
    \item \textbf{Chapters:} Specific, actionable investigation areas.
    \item \textbf{Queries:} 3-5 specific high-value search queries per chapter.
\end{enumerate}

\textbf{Output JSON Schema:}
\begin{verbatim}
{
    "root_task": "{root_topic}",
    "phases": [
        {
            "phase_title": "Phase 1: ...",
            "chapters": [
                {
                    "chapter_title": "...",
                    "search_queries": ["query1", "query2"]
                }
            ]
        }
    ]
}
\end{verbatim}
\end{reasoning_prompt}

\clearpage
\subsubsection{Researcher Agent}
\label{app:researcher}

\begin{reasoning_prompt}{System Prompt: Researcher Agent (Sub-task Execution)}
\textbf{Role:} Senior Research Analyst (Deep Research Agent)\\
\textbf{Mission:} Conduct a rigorous investigation on the specific chapter below.\\
\textbf{Target:}
\begin{itemize}
    \item \textbf{Root Topic:} \{root\_topic\}
    \item \textbf{Current Phase:} \{task\_obj.phase\_title\}
    \item \textbf{Current Chapter:} \{task\_obj.chapter\_title\}
    \item \textbf{Search Strategy:} \{task\_obj.question\}
\end{itemize}

\textbf{Context Awareness (DAG Memory):}
The following findings have already been established in previous chapters.
\textbf{CRITICAL: Do NOT repeat general facts found below. Build upon them. Focus on "What's missing" or "Next order effects".}\\
\{previous\_context if previous\_context else "[No previous context]"\}

\textbf{Output Requirements (Professional Standard):}
\begin{enumerate}
    \item \textbf{Format:} Markdown.
    \item \textbf{Visuals, Formatting \& References:}
    \begin{itemize}
        \item \textbf{Markdown Tables \& Figures:} Must have numbered captions. Cell content must be CONCISE.
        \item \textbf{LaTeX Formulas:} Use standard LaTeX syntax. Append tags/IDs to important equations.
        \item \textbf{Mermaid Diagrams:} Strictly use \texttt{graph LR} or \texttt{gantt}.
    \end{itemize}
\end{enumerate}
\end{reasoning_prompt}

\subsubsection{Summarizer Agent}
\label{app:summarizer}

\begin{reasoning_prompt}{System Prompt: Summarizer Agent (Context Transition)}
\textbf{Task:} Summarize Section Context\\
\textbf{Input Text:} \{input\_text\}

\textbf{Goal:}
Write a concise summary (2-3 sentences) of the key findings and arguments in this section.
This summary will be used by the writer of the NEXT section to ensure a smooth logical transition.
\end{reasoning_prompt}

\subsubsection{Graph Construction Methodology}
\label{app:graph_gen}

The following prompts illustrate the pipeline for transforming unstructured text into a structured Research Graph.

\begin{reasoning_prompt}{Step 1: Construct Insight (Atomic Extraction)}
\textbf{Task:} Identify Analytical Insights\\
\textbf{Input Sentences:} \{batch\_json\}\\
\textbf{Expert Advice:} \{\}

\textbf{Instruction:}
Following the Expert Advice, Review the sentences above from a research report.
\textbf{Select ONLY} the sentences that represent \textbf{Analysis, Synthesis, Deductions, or Conclusions}.
\begin{itemize}
    \item \textbf{REJECT:} Pure descriptions of data, procedural steps, or obvious definitions.
    \item \textbf{KEEP:} Sentences that connect dots, explain "why", or summarize trends/implications.
\end{itemize}

\textbf{Output JSON Schema:}
\begin{verbatim}
{ "selected": ["Exact text of insight 1", "Exact text of insight 2"] }
\end{verbatim}
\end{reasoning_prompt}

\clearpage
\begin{reasoning_prompt}{Step 2: Extract Residual Context (Missing Info)}
\textbf{Task:} Extract Residual Useful Information from Research Report\\
\textbf{Original Report Snippet:} \{task['report\_text']\}\\
\textbf{Context:} We have already extracted \{facts\_count\} atomic facts/insights from this report into a database.

\textbf{Goal:} Capture the \textbf{Missing Information Context}. What important context, statistics, table, figure, latex or descriptive details are lost but is useful if we only keep the atomic facts/insights?

\textbf{Instruction:}
Write a dense summary (1000 words) of the "Residual Useful Information". Focus on:
\begin{enumerate}
    \item Figures or Tables or latex formulas which keep dense information.
    \item Implicit tensions or qualitative descriptions not easily captured as "facts".
    \item The evolution of the situation.
\end{enumerate}

\textbf{Output JSON:} \texttt{\{ "dense\_summary": "..." \}}
\end{reasoning_prompt}

\begin{reasoning_prompt}{Step 3: Construct Links (Logical Entailment)}
\textbf{Task:} Build Logical Entailment Links\\
\textbf{Scenario:} We are analyzing the thought process at \textbf{Step \{child\_id\}}. This step logically follows/relies on \textbf{Step(s) \{parent\_ids\}}.

\textbf{Nodes in Current Step (Findings):} \{c\_json\}\\
\textbf{Nodes in Previous Step (Evidence):} \{p\_json\}

\textbf{Instructions:}
Identify logical connections (\texttt{source} - \texttt{target}).
\textbf{Allowed Link Types:}
\begin{enumerate}
    \item \textbf{Internal Link (Self-Consistency):} Source and Target are BOTH in Current Step.
    \item \textbf{Cross-Layer Link (Dependency):} Source is from Previous Step, Target is in Current Step.
\end{enumerate}
\textbf{Constraint:} Arrows generally flow forward in logic (Evidence - Conclusion) or forward in time.

\textbf{Output JSON Schema:}
\begin{verbatim}
{
    "links": [
        { "source": "insight_1", "target": "insight_2", "relation": "inference" },
        { "source": "fact_1", "target": "insight_5", "relation": "supports" }
    ]
}
\end{verbatim}
\end{reasoning_prompt}

\clearpage
\begin{reasoning_prompt}{Step 4: Construct Global Insight (Multi-Source Synthesis)}
\textbf{Role:} Chief Research Scientist (Meta-Synthesis Architecture)\\
\textbf{Task:} Multi-Source Pattern Recognition \& Iterative Insight Generation\\
\textbf{Input Data:} Compilation of knowledge nodes (atomic facts OR high-level insights).

\textbf{Instructions:}
\begin{enumerate}
    \item \textbf{Cluster \& Converge:} Look for \textbf{clusters} of evidence where 3 or more disparate nodes point to a single, larger truth.
    \item \textbf{Iterative Synthesis:}
    \begin{itemize}
        \item Generate \textbf{5-15 Global Insights}.
        \item Combine existing Insights to form a "Higher-Order Meta-Insight".
        \item Each Global Insight must be a novel synthesis, not regurgitation.
    \end{itemize}
    \item \textbf{Reasoning Trace:} Articulate \textit{how} these scattered nodes converge.
    \item \textbf{Strict Citation:} Every Global Insight must be supported by a list of specific \texttt{supporting\_node\_ids}.
\end{enumerate}

\textbf{Output JSON Schema:}
\begin{verbatim}
{
    "global_insights": [
        {
            "synthesis_logic": "We observe a convergence across [Node A]...",
            "content": "The finalized high-level conclusion...",
            "supporting_node_ids": ["chap_1_fact_5", "global_insight_2"]
        }
    ]
}
\end{verbatim}
\end{reasoning_prompt}

\subsubsection{Writer Agent}
\label{app:writer}

\begin{reasoning_prompt}{System Prompt: Writer Agent (Report Generation)}
\textbf{Role:} Senior Strategic Analyst (Think Tank Level)\\
\textbf{Task:} Write Chapter \textbf{\{title\}}\\
\textbf{Input Data:} Verified Quiz Facts, Knowledge Graph Nodes, and Narrative Context.

\textbf{Writing Guidelines:}
\begin{enumerate}
    \item \textbf{Flow:} \{flow\_context\}
    \item \textbf{Deep \& Dense:} Avoid "AI-style" lists. Write comprehensive paragraphs.
    \item \textbf{Header Formatting:} Identify Section Number. Use \texttt{\#\#\#} (H3) for subsections.
    \item \textbf{Visuals:}
    \begin{itemize}
        \item Use Python Plotting Agent (SciencePlots/Seaborn) for statistical data.
        \item Use LaTeX for formulas with continuous numbering tags.
    \end{itemize}
\end{enumerate}

\textbf{Negative Constraints:}
\begin{itemize}
    \item NO CITATIONS IN DRAFT (Tags are added later).
    \item NO CHINESE.
    \item NO BROKEN TABLES.
\end{itemize}
\end{reasoning_prompt}

\clearpage

\subsubsection{Automated Evaluation Metric Construction}
\label{app:metric}

\begin{reasoning_prompt}{Prompt: Exam QA Generation (Quiz)}
\textbf{Task:} Quiz Generation\\
\textbf{Requirements:}
\begin{enumerate}
    \item \textbf{Quantity \& Format:} Generate exactly one question per step (MCQ, True/False, Fill-in-Blank).
    \item \textbf{Self-Contained Context (CRITICAL):} The testing LLM has NO access to the original report. Must embed all context (dates, names) in the question stem.
    \item \textbf{Depth Scoring (0-5):}
    \begin{itemize}
        \item 0-1: Surface-level fact.
        \item 4-5: Requires deep inference or cross-referencing.
    \end{itemize}
\end{enumerate}

\textbf{Output JSON Schema:}
\begin{verbatim}
{
    "quiz": [
        {
            "type": "multiple_choice",
            "question": "...",
            "options": ["A...", "B..."],
            "answer": "...",
            "depth_metric": 4
        }
    ]
}
\end{verbatim}
\end{reasoning_prompt}

\begin{reasoning_prompt}{Prompt: Dialectical Confidence Audit (Bias Calibration)}
\textbf{Task:} Dialectical Confidence Audit Generation\\
\textbf{Goal:} Identify topics characterized by \textbf{interpretative ambiguity} or \textbf{balanced conflict} ("Gray Zones").

\textbf{Rules for Formulation:}
\begin{enumerate}
    \item \textbf{Blind Audit Format:} Question must NOT reference report structure.
    \item \textbf{Dialectical Framing:} Use "Thesis vs. Antithesis" logic.
    \item \textbf{Hidden Consensus:} Test if respondent can identify inconclusive data.
\end{enumerate}

\textbf{Output Format:}
\begin{verbatim}
{
    "question": "Regarding the net impact of 'Green Initiative'...",
    "options": [
        "Thesis: It is a success due to 30% reduction...",
        "Antithesis: Impact negated by 10% rise in total emissions..."
    ],
    "answer": [
        "Reason: Report confirms efficiency gains. Score: 6.",
        "Reason: Report acknowledges rebound effect. Score: 5."
    ],
    "type": "confidence_score"
}
\end{verbatim}
\end{reasoning_prompt}

\subsection{Task Selection}

The \textsc{SuperResearch} benchmark is meticulously constructed to challenge the limits of autonomous research agents. We selected 300 professional research tasks across 10 high-stakes domains, ensuring a balanced distribution of theoretical complexity and empirical depth. 

\begin{figure}[t]
    \centering
    \includegraphics[width=\linewidth]{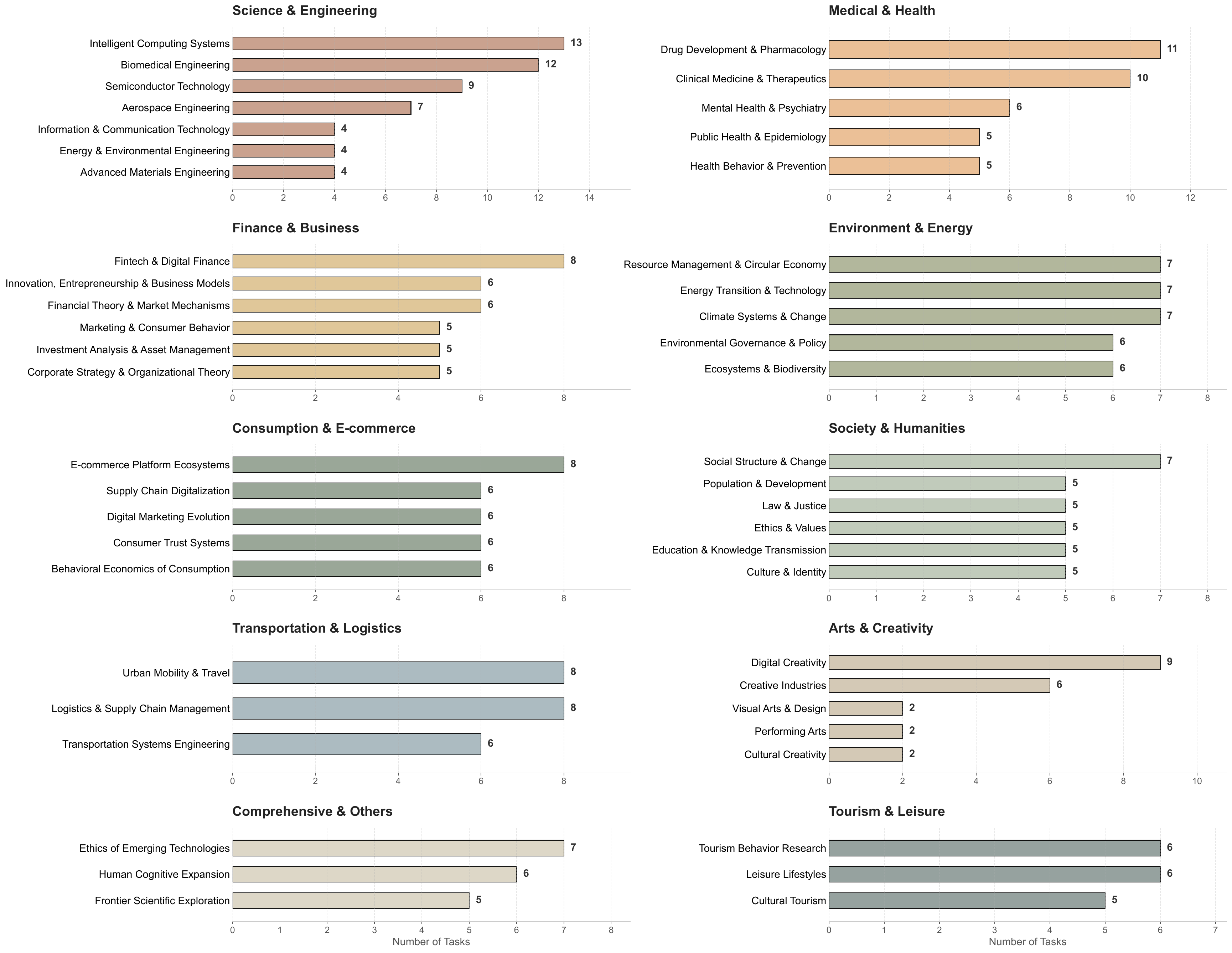}
    \vspace{-3mm}
    \caption{
        Detailed Task Distribution Across Subdomains. Supplementing the main text, this figure visualizes the granular distribution of tasks within each specific subdomain. This breakdown highlights the diversity of the \textsc{SuperResearch} benchmark, confirming its coverage across a wide spectrum of specialized professional fields.
    }
    \label{fig:data_distribution}
\end{figure}

As illustrated in Figure \ref{fig:data_distribution}, our dataset covers a wide spectrum of professional fields ranging from \textit{Science \& Engineering} and \textit{Medical \& Health} to specialized sectors like \textit{Environment \& Energy} and \textit{Finance \& Business}. The task selection process, spearheaded by a panel of Ph.D. candidates and industry experts, follows a "ceiling-level" philosophy: 
\begin{itemize}
    \item \textbf{Cognitive Nuance:} We explicitly prioritized tasks containing conflicting evidence or unresolved scientific tensions, requiring agents to perform dialectical reasoning.
    \item \textbf{Structural Depth:} Unlike traditional QA benchmarks, each task in \textsc{SuperResearch} is designed with long-horizon dependencies, where finding a definitive conclusion necessitates traversing multiple levels of intermediate evidence.
    \item \textbf{High Entropy Landscapes:} The information environment for these tasks is intentionally unstructured, forcing agents to demonstrate superior information filtering and synthesis capabilities.
\end{itemize}

\clearpage
\section{More Details about Evaluation Metrics.}

\begin{figure}[t]
    \centering
    \includegraphics[width=0.9\linewidth]{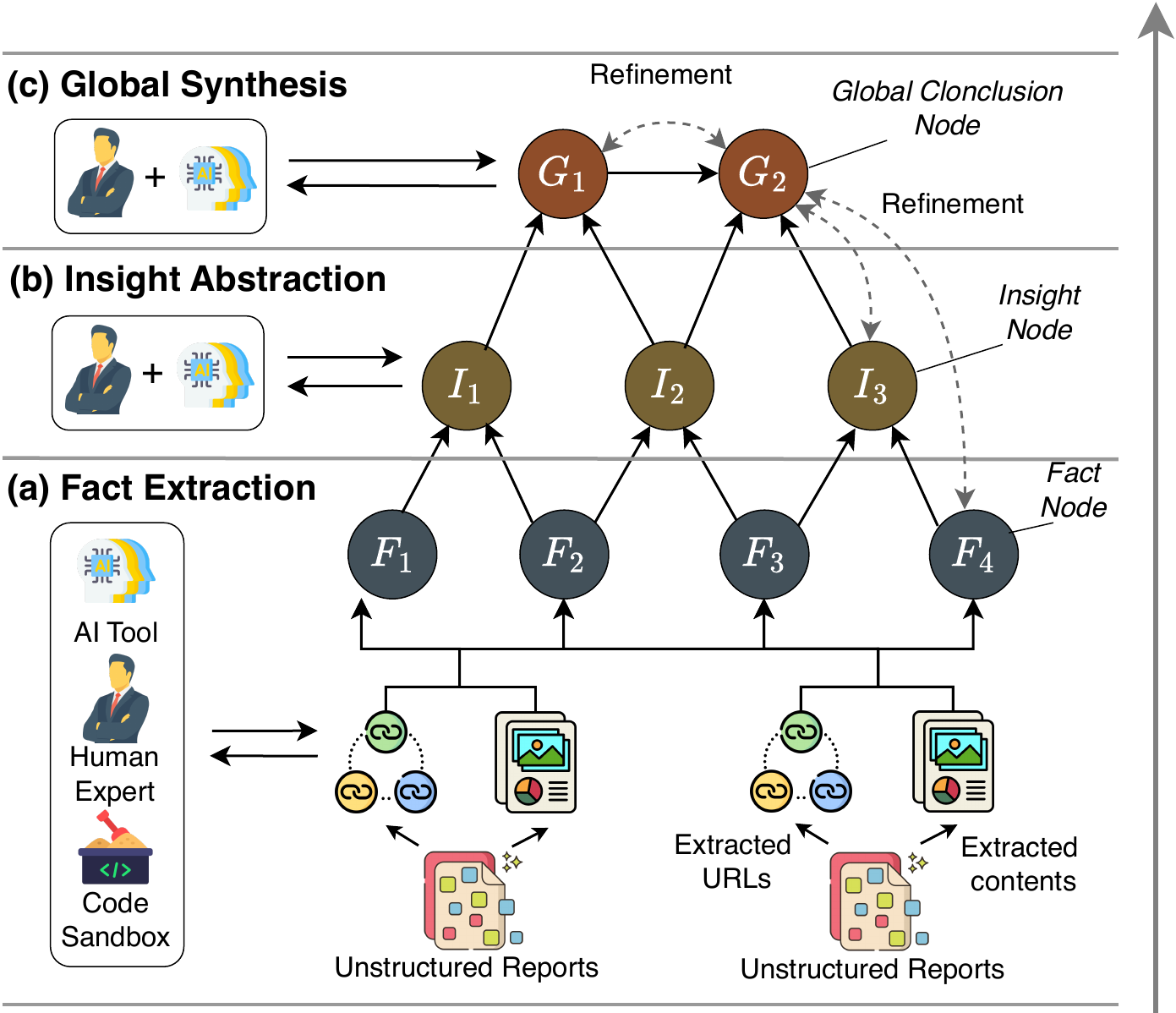} 
    \caption{Research Graph Construction Pipeline. The process transforms unstructured sub-reports into a structured knowledge graph in three stages: (a) \textbf{Fact Extraction}: Decomposing unstructured text into atomic fact nodes anchored to specific URLs; (b) \textbf{Insight Abstraction}: A collaborative human-AI process that derives higher-order reasoning nodes from fact clusters to build a bottom-up logic topology; (c) \textbf{Global Synthesis}: Merging disparate evidence clusters inter-connected global conclusions that serve as the ground truth.}
    \label{fig:graph_construction}
\end{figure}
\subsection{Coverage and Comprehension ($\mathcal{R}_{\text{weighted}}$)}\label{app:kc_formulation}
Unlike flat recall metrics, Coverage and Comprehension emphasize the retrieval of high-value structural information. We assign a topological weight $w(v)$ to each node $v$ in the Research Graph based on its hierarchical level (Atomic Fact, Key Insight, or Global Insight). Let $Children(v)$ be the set of nodes supporting $v$. The weight is defined recursively to reflect derivation depth:
\begin{equation}
    w(v) = 
    \begin{cases} 
    1 & \text{if } v \text{ is a leaf (Atomic Fact level)} \\
    1 + \max_{u \in Children(v)} w(u) & \text{otherwise (Insight levels)}
    \end{cases}
\end{equation}
The Depth-Weighted Recall is calculated as:
\begin{equation}
    \mathcal{R}_{\text{weighted}} = \frac{\sum_{v \in \mathcal{V}_{hit}} w(v)}{\sum_{v \in \mathcal{V}_{total}} w(v)} \times 100
\end{equation}
where $\mathcal{V}_{total}$ is the set of all graph nodes, and $\mathcal{V}_{hit}$ is the subset of nodes semantically verified as present in the generated report by an LLM judge.


\subsection{Logical Consistency ($\mathcal{C}_{\text{logic}}$)}\label{app:lc_formulation}
Logical Consistency evaluates the integrity of the reasoning chains. Let $\mathcal{G}$ be the set of Global Conclusion nodes (roots). A global node $g \in \mathcal{G}$ is considered \textit{supported} if and only if there exists a valid, unbroken citation path in the recovered subgraph from $g$ down to a recovered Atomic Fact node.
We formally define $\mathcal{C}_{\text{logic}}$ as the product of the valid inference ratio and overall graph coverage:
\begin{equation}
    \mathcal{C}_{\text{logic}} = \left( \frac{|\mathcal{G}_{supported}|}{|\mathcal{G}_{hit}| + \epsilon} \times \frac{|\mathcal{G}_{hit}|}{|\mathcal{G}_{total}|} \right) \times 100
\end{equation}
where $\epsilon$ is a small constant to prevent division by zero. This multiplicative formulation penalizes reports that hallucinate correct high-level answers without the necessary supporting evidence.


\subsection{Report Utility ($\mathcal{U}_{\text{qa}}$)}\label{app:qa_formulation}
We quantify the actionable value of the report using a set of $N$ exam questions $Q = \{q_1, ..., q_N\}$ derived from the Research Graph. Let $a_i$ be the model's answer to question $q_i$ generated in a closed-context setting (relying strictly on the report content), and $a_i^{\star}$ be the ground truth.
\begin{equation}
    \mathcal{U}_{\text{qa}} = \frac{1}{N} \sum_{i=1}^{N} \mathbb{I}(a_i \approx a_i^{\star}) \times 100
\end{equation}
where $\mathbb{I}(\cdot)$ is the semantic equivalence indicator function evaluated by the LLM judge.


\subsection{Objectivity Score ($\mathcal{O}_{\text{bias}}$)}\label{app:os_formulation}
To measure the Calibration of Stance, we define a Thesis $A$ and Antithesis $B$ for controversial topics. Let $S_A, S_B \in [0, 10]$ be the evidentiary support scores extracted from the generated report by a double-blind auditor, and $GT_A, GT_B$ be the ground-truth support levels derived from the graph. The calibration error $E$ is:
\begin{equation}
    E = |S_A - GT_A| + |S_B - GT_B|
\end{equation}
The final Objectivity Score is derived as:
\begin{equation}
    \mathcal{O}_{\text{bias}} = \max(0, 100 - E \times \lambda)
\end{equation}
In our experiments, we set the penalty factor $\lambda = 5$, ensuring that significant deviations in viewpoint balance heavily impact the score.


\subsection{Citation Health}\label{app:reliability_formulation}
Citation Health serves as a diagnostic suite to detect Single-Source Dependency. We compute two specific indicators based on the citation distribution $C$. Let $n$ be the total number of unique sources cited, and $p_i$ be the proportion of citations attributed to source $i$.

\paragraph{Source Dominance ($\mathcal{D}_{\text{src}}$).} This metric checks for volume imbalance. We utilize normalized Shannon Entropy ($H_{norm}$) to measure the "peakedness" of the citation distribution:
\begin{equation}
    H(C) = -\sum_{i=1}^{n} p_i \log_2 p_i, \quad \mathcal{D}_{\text{src}} = \left(1 - \frac{H(C)}{\log_2 n}\right) \times 100
\end{equation}
A high $\mathcal{D}_{\text{src}}$ indicates that the report relies disproportionately on a single source for its information volume.

\paragraph{Narrative Monopolization ($\mathcal{M}_{\text{mono}}$).} This metric evaluates the spatial layout of citations. Let $K$ be the total number of distinct sections in the report, and $k_i$ be the number of sections where source $i$ appears. We measure the maximum narrative coverage:
\begin{equation}
    \mathcal{M}_{\text{mono}} = \max_{i} \left( \frac{k_i}{K} \right) \times 100
\end{equation}
A high $\mathcal{M}_{\text{mono}}$ suggests that a single source drives the narrative flow across the entire document, rather than being synthesized with others.



\clearpage
\section{More Experiment Results.}

\subsection{Coverage and Comprehension Details}\label{app:coverage}
As shown in Table~\ref{tab:knowledge_capture}, Gemini Deep Research achieves the leading overall coverage through strong atomic and insight retrieval, whereas Native Search agents like Grok and Kimi-k2 demonstrate competitive capabilities in high-level synthesis.
\subsection{Sensitivity Analysis Details}\label{app:responsiveness}

\subsubsection{Responsiveness Rate Formulation}
To quantify the sensitivity of an evaluator, we define the \textbf{Responsiveness Rate (RR)} as the probability that the evaluator's score shifts in the correct direction following a controlled perturbation. 

Let $\mathcal{D} = \{r_1, r_2, ..., r_N\}$ be the set of original reports derived from $N$ experiment instances (where $N=120$ in our setup, comprising 12 models $\times$ 10 tasks). For each report $r_i$, we generate two perturbed variants: a degraded version $r_i^{-}$ (with fact removal) and an improved version $r_i^{+}$ (with fact injection).

Let $S(r)$ denote the score assigned by an evaluator to report $r$. We define the responsiveness indicator function $\mathbb{I}(\cdot)$ for the two scenarios:

\paragraph{Degradation Responsiveness ($RR_{deg}$).} 
In the degradation scenario, a responsive evaluator should penalize the quality drop, i.e., $S(r_i) > S(r_i^{-})$. The rate is calculated as:
\begin{equation}
    RR_{deg} = \frac{1}{N} \sum_{i=1}^{N} \mathbb{I}\left( S(r_i) - S(r_i^{-}) > \epsilon \right) \times 100\%
\end{equation}

\paragraph{Improvement Responsiveness ($RR_{imp}$).} 
In the improvement scenario, a responsive evaluator should reward the quality gain, i.e., $S(r_i^{+}) > S(r_i)$. The rate is calculated as:
\begin{equation}
    RR_{imp} = \frac{1}{N} \sum_{i=1}^{N} \mathbb{I}\left( S(r_i^{+}) - S(r_i) > \epsilon \right) \times 100\%
\end{equation}

where $\epsilon$ is a minimal threshold (set to 0 in our experiments) to strictly enforce directional correctness.

\subsubsection{Score Normalization and Comparison}
To ensure a fair comparison between the Graph Metric (scale 0-100) and the LLM Judge (typically scale 0-10), we normalized the scalar outputs during the delta calculation. Specifically, for the LLM Judge, we applied a scaling factor of $\alpha=10$ before computing the differentials: $\Delta_{LLM} = (S_{LLM}^{org} - S_{LLM}^{pert}) \times 10$.

This normalization ensures that the magnitude of score changes is comparable in the sensitivity distribution analysis, although the Responsiveness Rate itself is scale-invariant as it only measures the \textit{directionality} of the change.

\subsubsection{Perturbation Methodology}
The controlled perturbation was conducted via a semi-automated pipeline to ensure ground-truth validity:
\begin{enumerate}
    \item \textbf{Degradation (Fact Removal):} We explicitly removed 1$\sim$3 Atomic Facts (nodes in the research graph) or Logical Links from the report's reasoning section. The expected outcome is a decrease in the \textit{Breadth/Recall} or \textit{Depth/CoT} metric.
    \item \textbf{Improvement (Fact Injection):} We retrieved 1$\sim$3 high-relevance facts from the ground-truth Knowledge Graph that were originally missing from the report and injected them into the appropriate context. The expected outcome is an increase in the corresponding metric.
\end{enumerate}
\begin{table}[t] 
    \centering
    \scriptsize
    \caption{Coverage and Comprehension Breakdown.
    Detailed performance analysis of \textbf{Coverage \& Comprehension} across different knowledge granularities.
    Evaluation covers three architectural paradigms: \colorbox{blue!8}{Deep Research System}, \colorbox{orange!8}{Native Search-Integrated Agent}, and \colorbox{green!8}{Search-Augmented Baseline}.
    \textbf{Coverage} ($\mathcal{R}_{\text{weighted}}$) represents the overall score, calculated as the depth-weighted aggregation of Atomic Facts, Key Insights, and Global Conclusions.
    \best{Bold} and \secBest{underline} denote the best and second-best performance, respectively; $\uparrow$ indicates that higher values are preferred.
    }
    \renewcommand{\arraystretch}{1.2} 
    \setlength{\tabcolsep}{10pt}
    
    \begin{tabular}{l cccc}
        \hline
        \rowcolor{gray!6} & \multicolumn{3}{c}{\textbf{Knowledge Recall Breakdown}} & \\
        \cline{2-4}
        \rowcolor{gray!6} \multirow{-2}{*}{\textbf{Method}} 
        & \textbf{Atomic Facts}~$\uparrow$ 
        & \textbf{Key Insights}~$\uparrow$ 
        & \textbf{Global Conclusions}~$\uparrow$ 
        & \multirow{-2}{*}{\makecell{\textbf{Overall} \\ \textbf{Coverage} ($\mathcal{R}_{\text{weighted}}$)}~$\uparrow$} \\
        \hline

        \rowcolor{blue!8} \multicolumn{5}{c}{Deep Research System} \\
        \hline
        Gemini Deep Research & \best{27.94} & \best{38.56} & \secBest{22.70} & \best{33.15} \\
        Sonar Deep Research & 23.69 & 32.48 & 16.67 & 28.26 \\
        Tongyi Deep Research & 21.69 & 33.63 & 19.58 & 27.41 \\
        o3 Deep Research & 19.75 & 27.89 & 19.65 & 23.44 \\
        o4 Mini Deep Research & 19.22 & 27.71 & 19.77 & 23.11 \\
        \hline

        \rowcolor{orange!8} \multicolumn{5}{c}{Native Search-Integrated Agent} \\
        \hline
        Kimi-k2 & 20.69 & 31.67 & \best{24.58} & 25.97 \\
        Grok-4-1-fast & \secBest{26.35} & \secBest{34.94} & 19.23 & \secBest{30.09} \\
        \hline
        
        \rowcolor{green!8} \multicolumn{5}{c}{Search-Augmented Baseline} \\
        \hline
        Deepseek-r1 & 19.67 & 28.65 & 15.45 & 24.21 \\
        Claude-3.5-sonnet & 24.62 & 31.21 & 17.30 & 27.02 \\
        Minimax-m2.1 & 17.06 & 25.37 & 12.89 & 20.91 \\
        Llama-3.3-70B & 14.03 & 17.97 & 13.15 & 15.55 \\
        Qwen-2.5-72B & 9.12 & 12.51 & 9.90 & 10.47 \\

        \hline
    \end{tabular}
    \label{tab:knowledge_capture}
\end{table}
\begin{figure*}[t]
    \centering
    \includegraphics[width=0.95\linewidth]{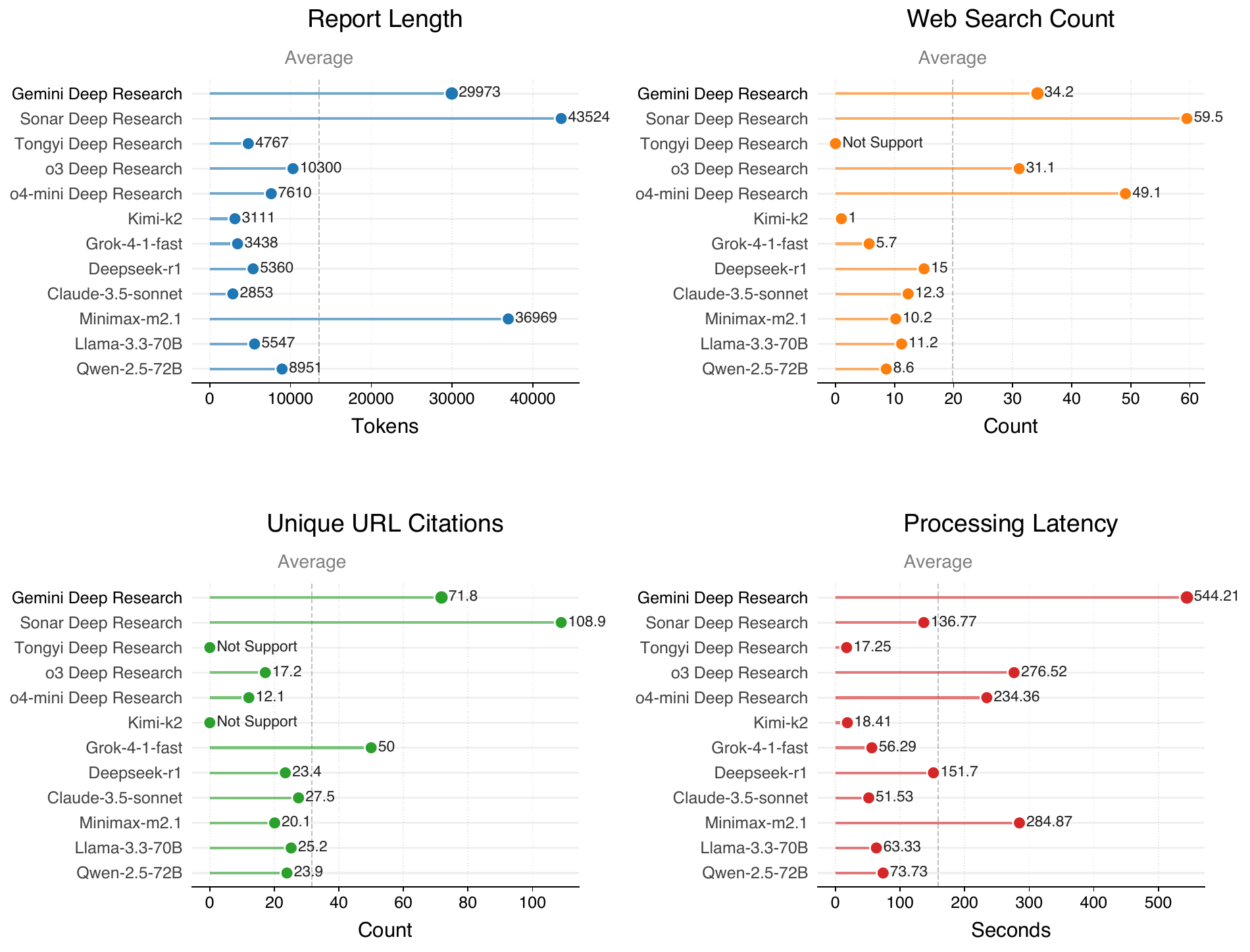}
    \caption{
        Multi-dimensional Operational Benchmarking across State-of-the-Art LLMs.
        We compare 12 models on our expert-curated benchmark. Each sub-panel represents a key operational metric, with the vertical dashed line indicating the industry average. Gemini Deep Research (highlighted) consistently demonstrates a superior balance between investigation depth (search counts) and synthesis volume (tokens), while maintaining high sourcing diversity. Models with zero values in specific search/source metrics are marked as Not Support.
    }
    \label{fig:model_metrics_benchmarking}
\end{figure*}

\subsection{Cross-Domain Performance Analysis}
To assess the robustness of Super Research across varying professional contexts, we evaluate model performance across 10 distinct domains, ranging from technical fields like \textit{Science \& Engineering} and \textit{Medical \& Health} to social and creative dimensions such as \textit{Society \& Humanities} and \textit{Arts \& Creativity}.

\begin{figure}[t]
    \centering
    \includegraphics[width=\linewidth]{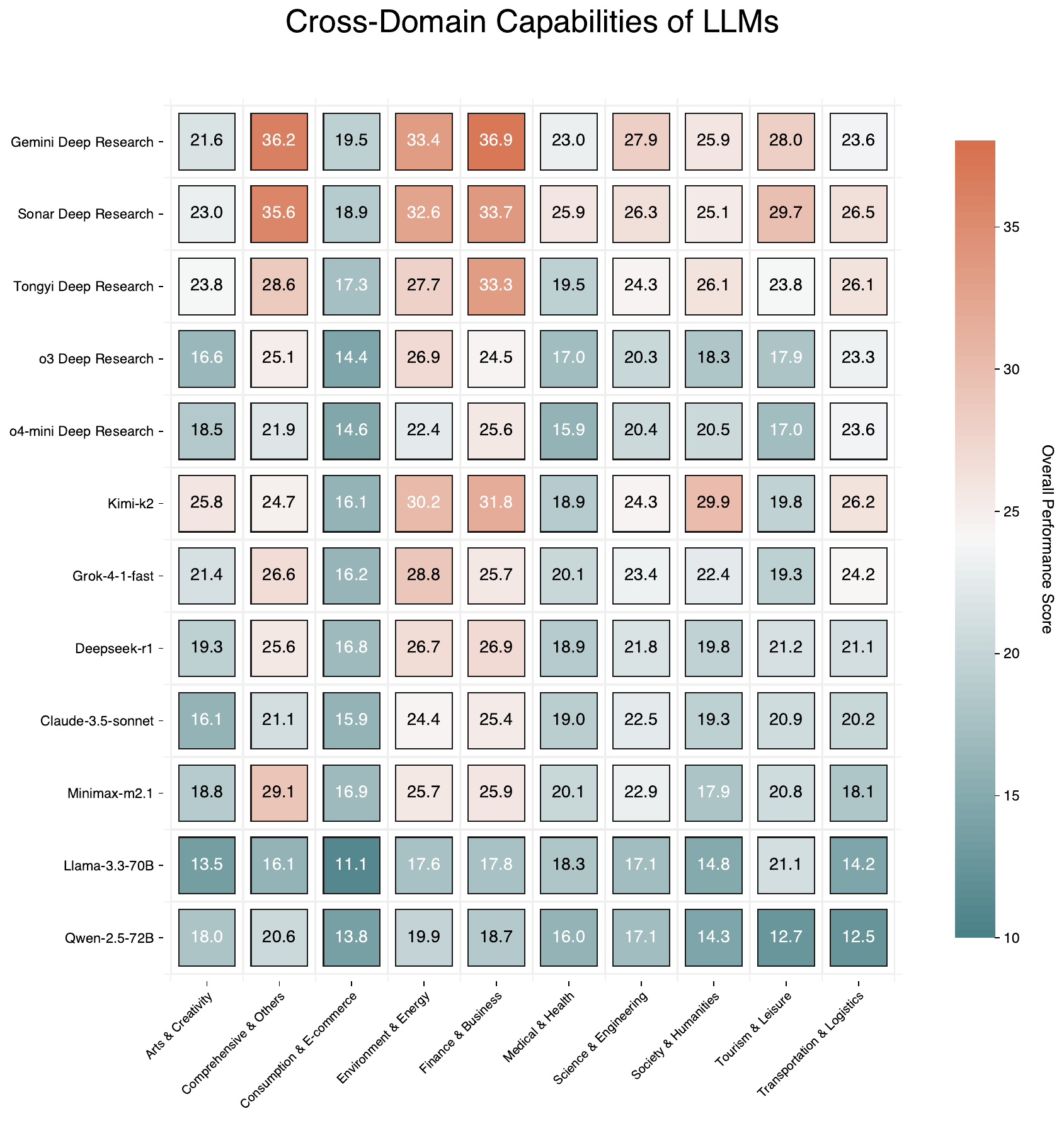}
    \vspace{-3mm}
    \caption{
        Cross-Domain Capabilities of LLMs. 
        Performance comparison across 10 specialized domains. The radar chart illustrates the overall capability score for each model tier. Gemini Deep Research shows particular dominance in business-analytical and technical-heavy sectors.
    }
    \label{fig:cross_domain_capabilities}
\end{figure}

The vertical-specific analysis in Figure \ref{fig:cross_domain_capabilities} reveals that Gemini Deep Research achieves its highest performance in \textit{Finance \& Business} and \textit{Environment \& Energy}, where structured decomposition and multi-layered verification are critical for navigating complex regulatory and technical documents. In contrast, while models like Sonar Deep Research show competitive results in \textit{Medical \& Health} and \textit{Tourism \& Leisure}, our framework maintains a more balanced profile across all 10 indices. These results suggest that the Super Research paradigm is not only deeper but also more versatile, effectively bridging the gap between niche-specific knowledge and general investigation capabilities.

\clearpage
\subsection{Ablation Study Prompts}
\label{app:ablation_prompts}

To validate the responsiveness of our metrics, we employed controlled perturbation experiments. The following prompts were used to artificially degrade or improve report quality, as well as to conduct the baseline "LLM Judge" evaluation.

\begin{reasoning_prompt}{Prompt: Report Degradation (Malicious Editor)}
\label{prompt:ablation_degrade}
\textbf{System Prompt:} You are a malicious editor. Your goal is to degrade the report quality.\\
\textbf{User Prompt:}
Original Report: \{report\_text\}

\textbf{Task:} REWRITE the report to specifically \textbf{REMOVE 1$\sim$3 facts of evidence}.
\begin{itemize}
    \item The report must remain readable and coherent, but these specific points must be gone.
    \item For sections not directly related to Facts, please copy them in full, and only make surgical adjustments to the relevant parts.
\end{itemize}

\textbf{Output:} ONLY the rewritten report text.
\end{reasoning_prompt}

\begin{reasoning_prompt}{Prompt: Report Improvement (Helpful Editor)}
\label{prompt:ablation_improve}
\textbf{System Prompt:} You are a helpful editor. Your goal is to improve the report.\\
\textbf{User Prompt:}
Original Report: \{report\_text\}

\textbf{Task:} REWRITE the report to \textbf{INSERT the following missing facts}. Integrate them naturally into relevant sections.

\textbf{Facts to INSERT:} \{facts\_list\}

\begin{itemize}
    \item For sections not directly related to Facts, please copy them in full, and only make surgical adjustments to the relevant parts.
\end{itemize}

\textbf{Output:} ONLY the rewritten report text.
\end{reasoning_prompt}

\begin{reasoning_prompt}{Prompt: LLM Direct Judge (Baseline Evaluator)}
\label{prompt:llm_judge}
\textbf{Role:} Expert Research Evaluator (Senior Research Analyst)\\
\textbf{User Question:} \{root\_question\}\\
\textbf{Report Content:} \{report\_text\}...

\textbf{Task:} Rate this report on two dimensions (0-10 scale):
\begin{enumerate}
    \item \textbf{Breadth (Recall):} How comprehensively does it cover the necessary key facts?
    \item \textbf{Depth (CoT/Logic):} How well does it explain the logical connections and underlying mechanisms?
\end{enumerate}

\textbf{Output JSON:}
\begin{verbatim}
{
    "breadth_score": float,
    "depth_score": float,
    "reason": "string"
}
\end{verbatim}
\end{reasoning_prompt}

\clearpage
\begin{reasoning_prompt}{Prompt: Comprehensive Quality Evaluation (5-Metric Baseline)}
\label{prompt:comprehensive_eval}
\textbf{Role:} Expert Research Evaluator\\
\textbf{Task:} Evaluate the quality of a research report generated in response to a specific user query.

\textbf{Input Report:} \{report\}

\textbf{Evaluation Metrics (Scale 0.0 - 10.0):}
\begin{enumerate}
    \item \textbf{Depth:} Does it provide deep technical insights and causal analysis?
    \item \textbf{Breadth:} Does it cover all necessary aspects and sub-topics?
    \item \textbf{Utility:} Is it practically useful and directly answers the intent?
    \item \textbf{Citation Quality:} Are citations frequent, relevant, and well-distributed?
    \item \textbf{Objective Score:} Does it provide multiple perspectives on controversial issues?
\end{enumerate}

\textbf{Output JSON:}
\begin{verbatim}
{
    "depth_score": <float>,
    "breadth_score": <float>,
    "utility_score": <float>,
    "citation_quality_score": <float>,
    "objective_score": <float>
}
\end{verbatim}
\end{reasoning_prompt}
\clearpage
\section{Case Study}
\label{sec:appendix_cases}

In this section, we provide detailed qualitative analyses of representative failure modes identified in our main results. These cases illustrate the specific behavioral patterns underpinning the quantitative metrics.

\subsection{Case Study: The Complexity Trap and Logic Bottleneck}
\label{app:case_failure_lack_consistency}
\begin{figure*}[h]
    \centering
    \includegraphics[width=0.95\linewidth]{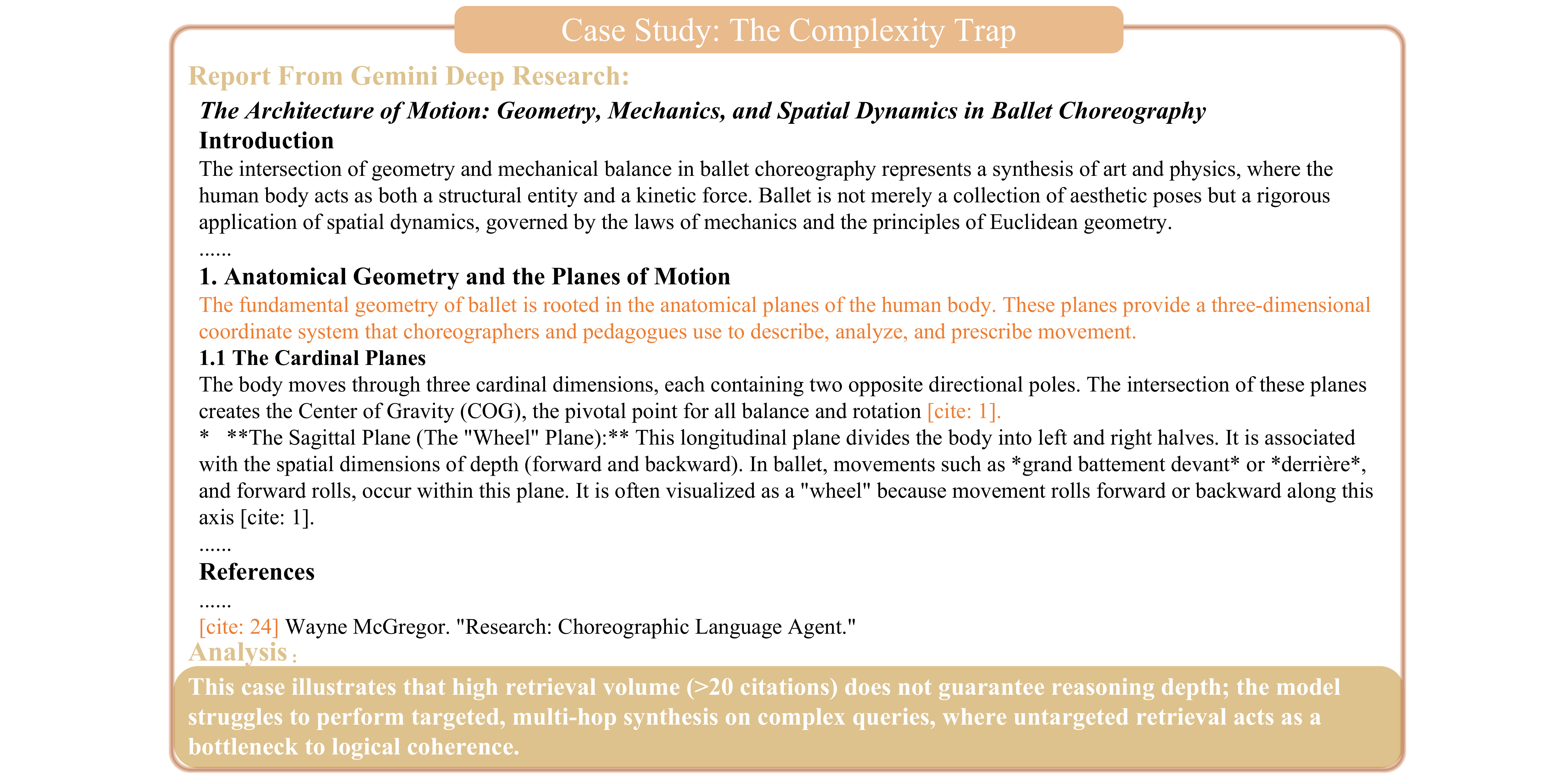}
    \caption{
        Case Study: The Complexity Trap
    }
    \label{fig:case_complexity}
\end{figure*}

Figure~\ref{fig:case_complexity} presents a case study illustrating how models struggle with the complexity of such queries, leading to significant logic bottlenecks.\\
\textbf{Query:} ``How are geometry and mechanical balance explored through spatial dynamics in ballet choreography?''


\subsection{Case Study: Single-Source Regression}
\label{app:case_overreliance}
\begin{figure*}[h]
    \centering
    \includegraphics[width=0.95\linewidth]{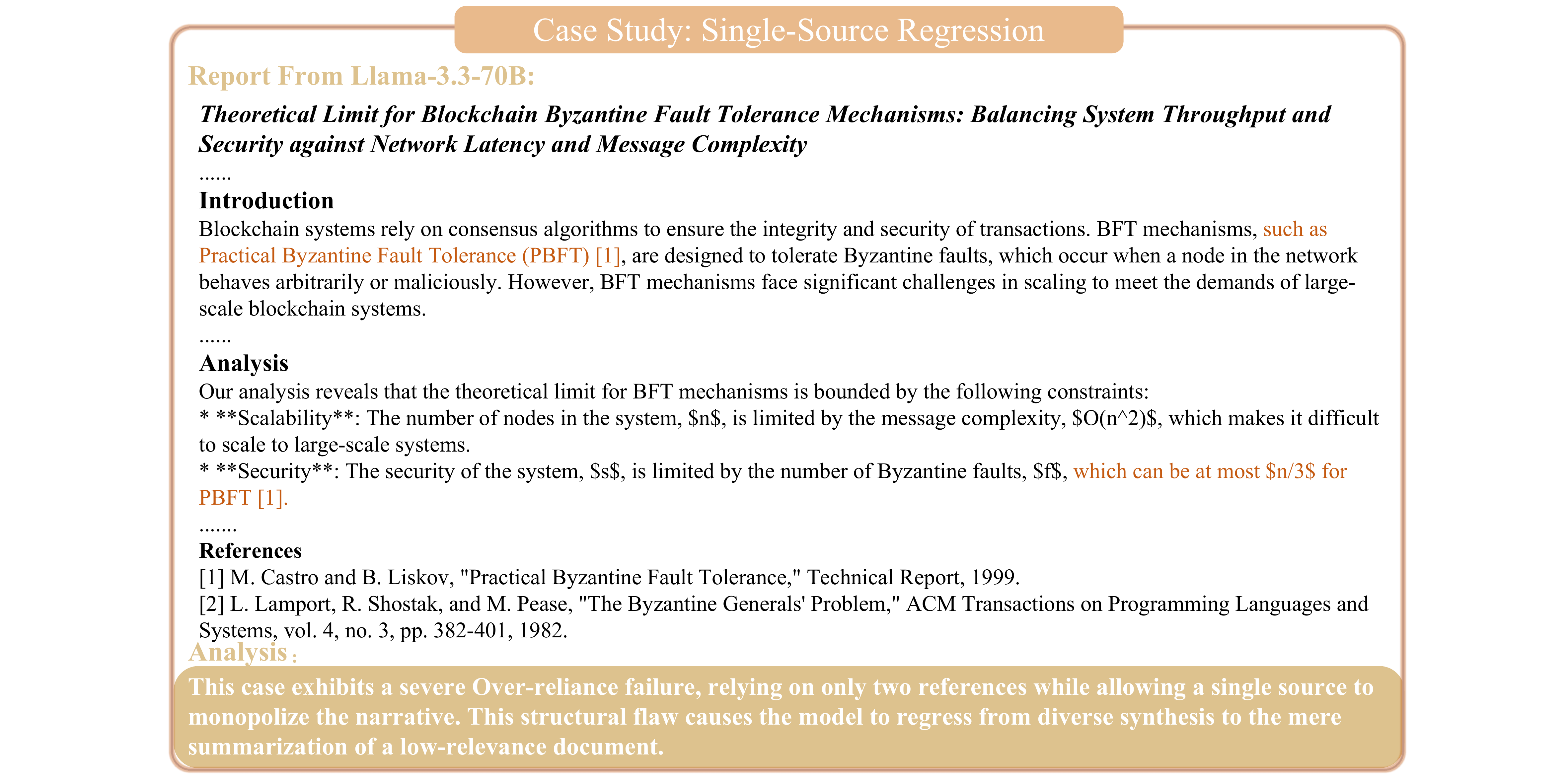}
    \caption{
    Case Study: Single-Source Regression
    }
    \label{fig:case_single}
\end{figure*}

Figure~\ref{fig:case_single} illustrates a typical instance of single-source regression, where the model's narrative is monopolized by a limited set of references despite the query's complexity.
\textbf{Query:} ``What represents the theoretical limit for blockchain Byzantine Fault Tolerance mechanisms when balancing system throughput and security against network latency and message complexity?''

\subsection{Case Study: The Illusion of Diversity via Spatial Overload}
\label{app:case_overload}
\begin{figure*}[h]
    \centering
    \includegraphics[width=0.95\linewidth]{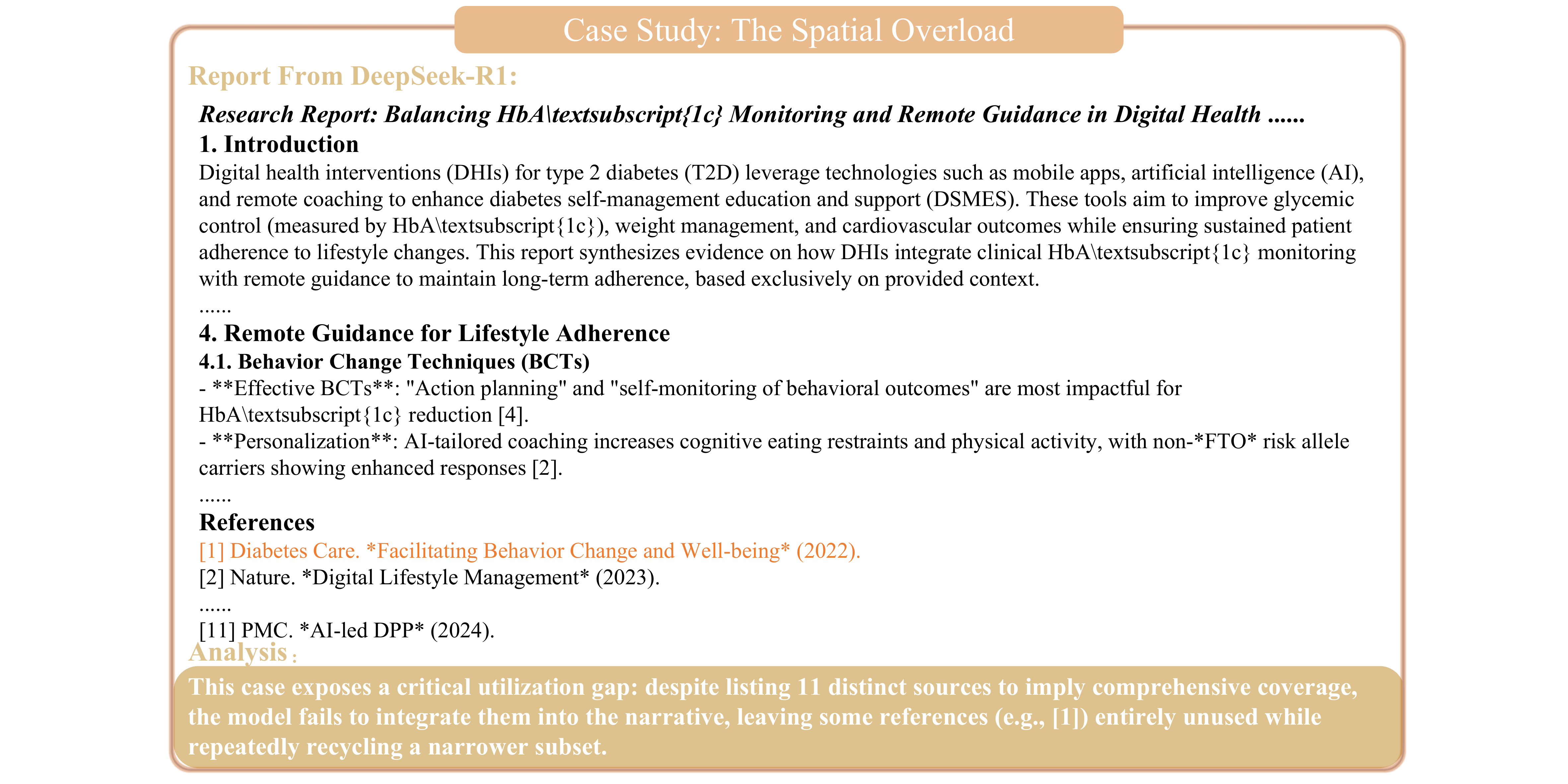}
    \caption{
    Case Study: The Spatial Overload
    }
    \label{fig:case_overload}
\end{figure*}

Figure~\ref{fig:case_overload} depicts a critical utilization gap, where the model lists numerous sources to imply comprehensive coverage but fails to integrate them into the narrative, effectively creating an illusion of diversity. 
\textbf{Query:} ``How do digital health interventions for type 2 diabetes balance clinical HbA1c monitoring with remote guidance to maintain adherence to lifestyle changes?''

\subsection{Case Study: Reliability Paradox via Defensive Summarization}
\label{app:case_conservatism}
\begin{figure*}[h]
    \centering
    \includegraphics[width=0.95\linewidth]{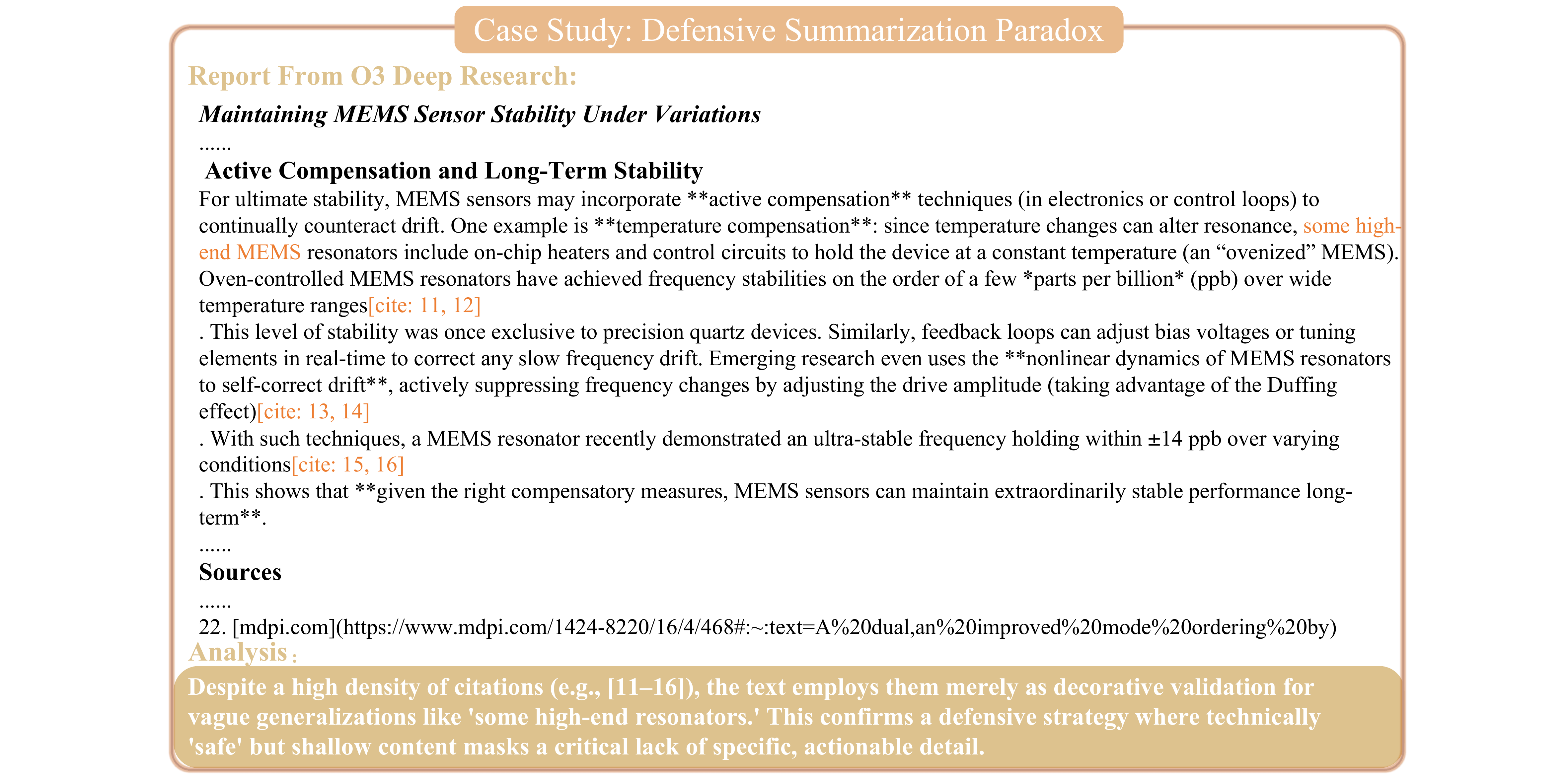}
    \caption{
    Case Study: Defensive Summarization Paradox
    }
    \label{fig:case_defensive}
\end{figure*}

Figure~\ref{fig:case_defensive} highlights the defensive summarization paradox, where the model employs a high density of citations to validate vague generalizations, resulting in technically `safe' but shallow content that lacks specific, actionable detail.
\textbf{Query:} ``Is it possible for MEMS sensors to maintain long-term resonant frequency stability and detection accuracy despite the combined effects of process deviations and environmental vibration?''
\clearpage
\section{Additional Materials}

\subsection{Web-based Analysis Tool}
\label{app:web_tool}

\begin{figure*}[t]
    \centering
    \includegraphics[width=0.95\linewidth]{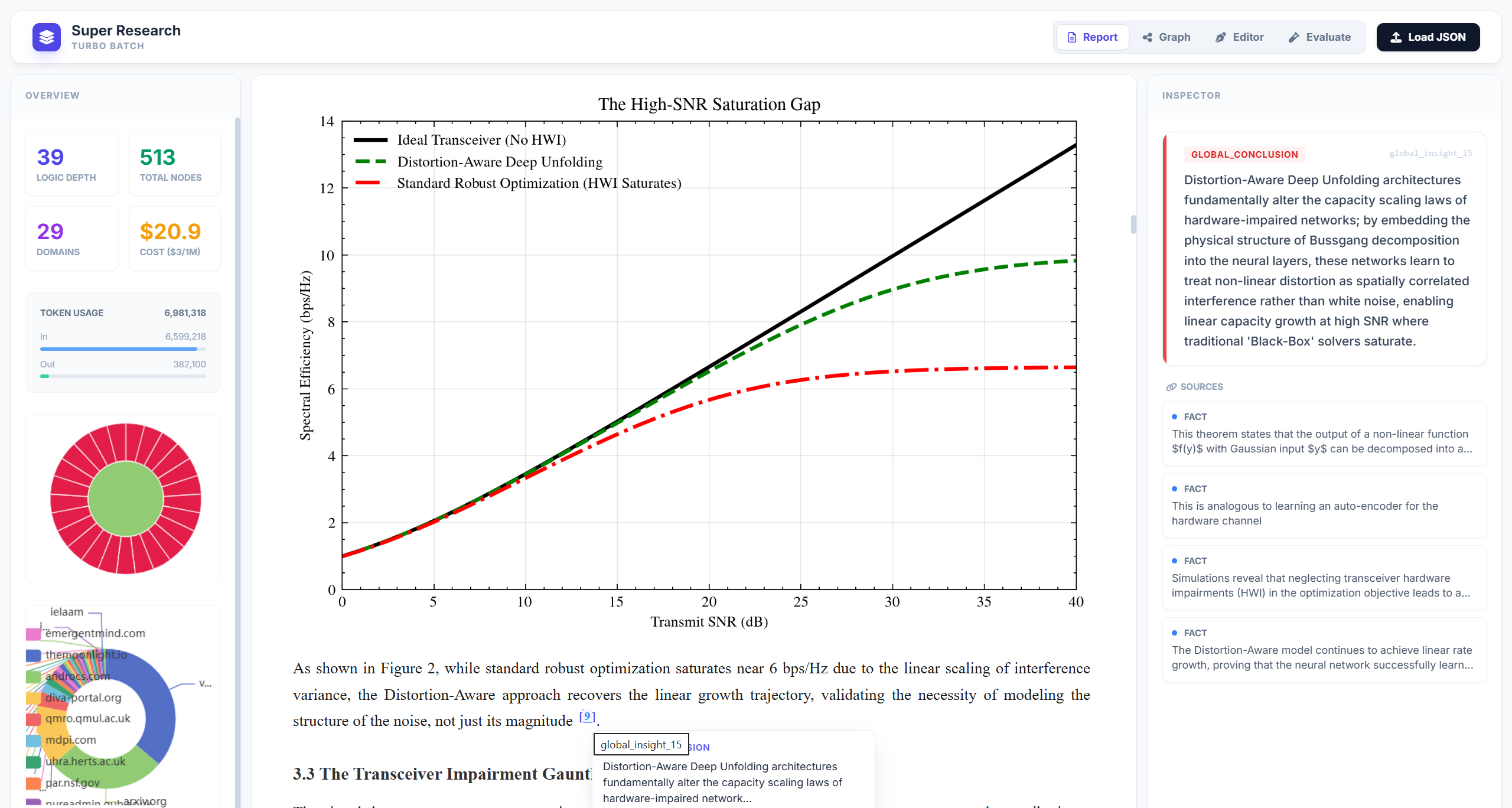}
    \caption{
        Report Management and Visualization Interface.
        (1) \textbf{Overview Dashboard}: The left sidebar tracks high-level metrics, including Logic Depth (177), Total Nodes (551), and estimated cost. 
        (2) \textbf{Interactive Rendering}: The main view displays the generated "Analyst-Grade" report. Note the high density of clickable citations (e.g., [1][2]...), which allow users to trace claims back to their source evidence.
    }
    \label{fig:web_report}
\end{figure*}

\begin{figure*}[t]
    \centering
    \includegraphics[width=0.95\linewidth]{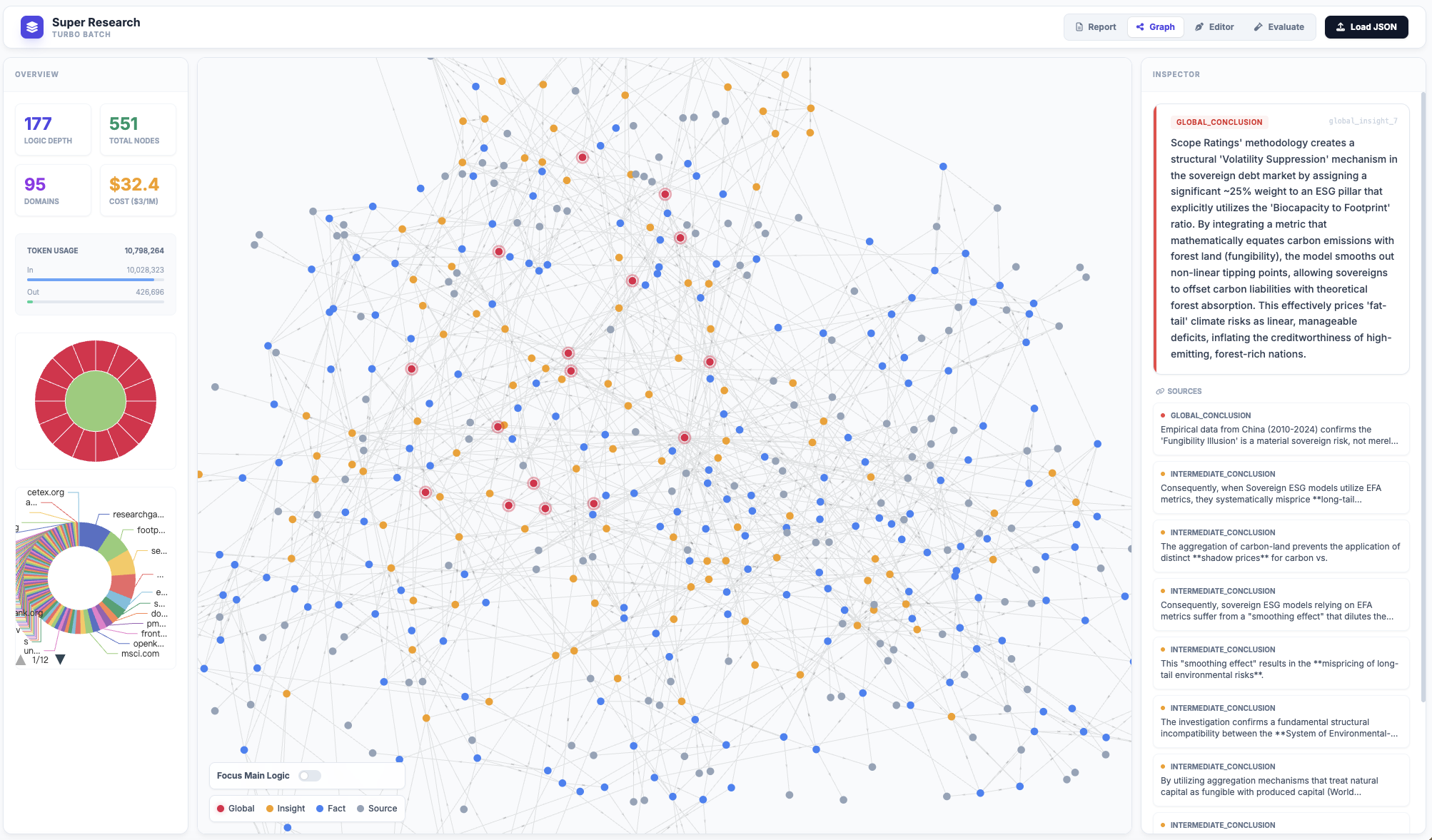}
    \caption{
        Interactive Research Graph Visualization.
        This interface projects the reasoning logic into a structured topology. 
        (1) \textbf{Hierarchical Nodes}: Nodes are color-coded by abstraction level: \textcolor{red}{Red} (Global Conclusions), \textcolor{orange}{Orange} (Intermediate Insights), and \textcolor{blue}{Blue} (Atomic Facts). 
        (2) \textbf{Inspector Panel}: The right sidebar details the specific textual content and source linkage for any selected node (e.g., "GLOBAL\_CONCLUSION"), exposing the semantic connections.
    }
    \label{fig:web_graph}
\end{figure*}

\begin{figure*}[t]
    \centering
    \includegraphics[width=0.95\linewidth]{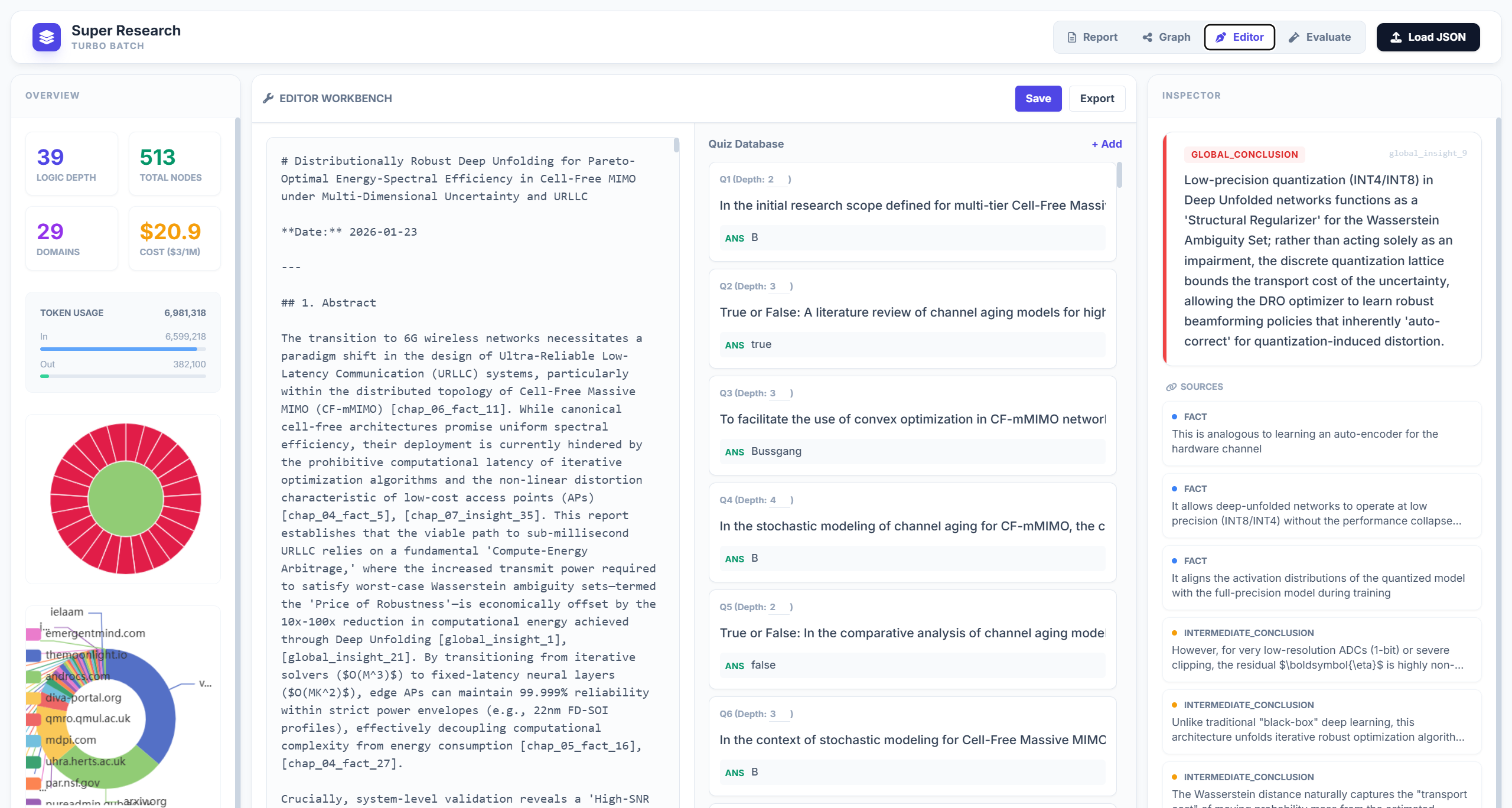}
    \caption{
        Split-View Evaluation Editor.
        To ensure the validity of automated metrics, this interface couples the report content with the evaluation set. 
        (1) \textbf{Editor Workbench}: The left panel displays the raw Markdown of the report. 
        (2) \textbf{Quiz Database}: The right panel lists generated exam questions (e.g., Q1, Q2), tagged by "Depth". Domain experts can use this view to intervene, editing or adding questions to correct alignment errors before the final audit.
    }
    \label{fig:web_question}
\end{figure*}

\begin{figure*}[t]
    \centering
    \includegraphics[width=0.95\linewidth]{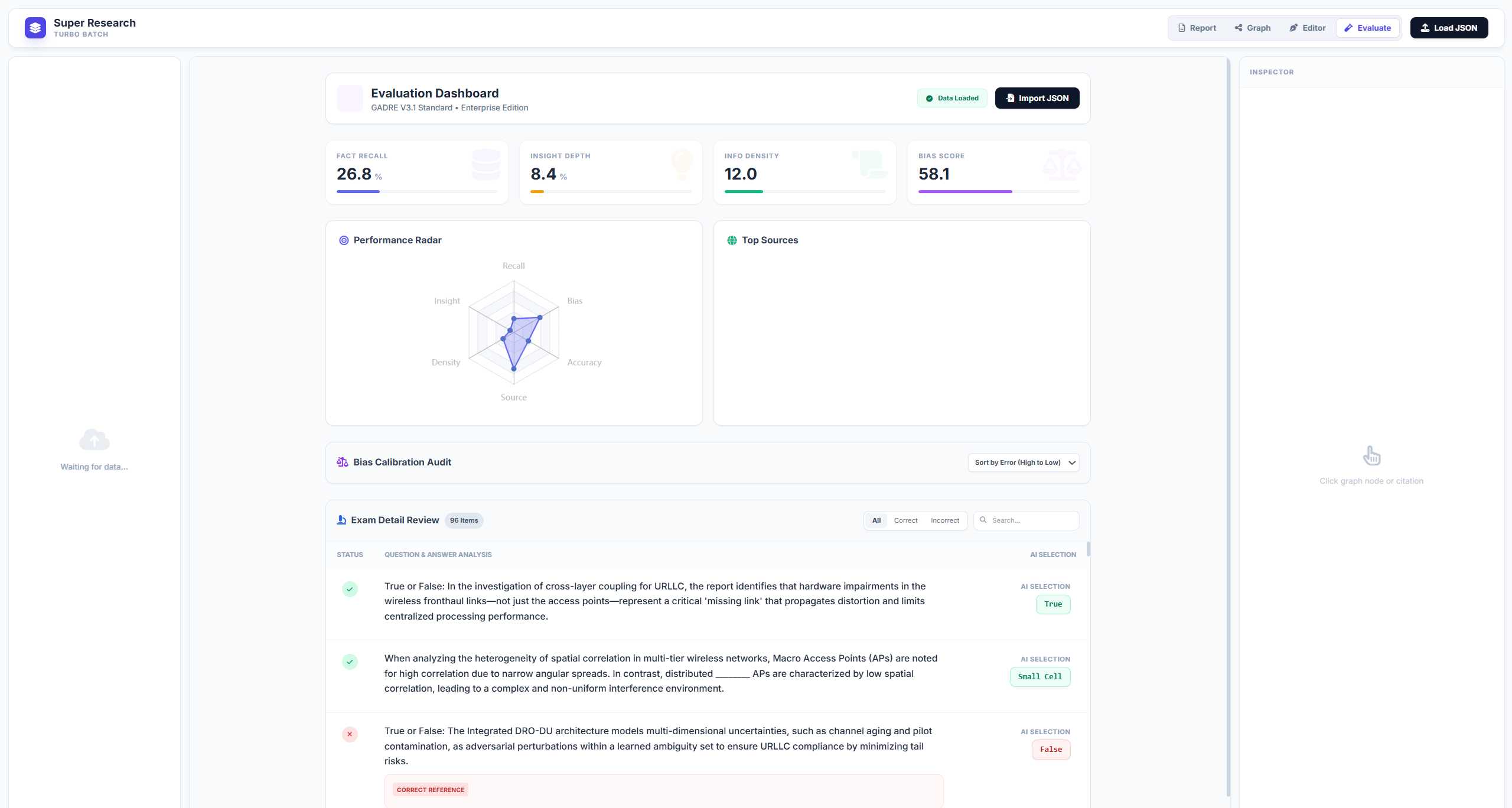}
    \caption{
        Visualization of the Evaluation Interface. This figure displays a comprehensive breakdown of metric scores, a radar chart for holistic comparison, and detailed performance results from the Exam QA.
    }
    \label{fig:web_eval}
\end{figure*}

To facilitate rigorous inspection and enable human-in-the-loop verification, we have developed a dedicated web-based platform for \textbf{Super Research}. The interface supports the full evaluation lifecycle, guiding users through four distinct stages: 
(1) Report Visualization (Figure~\ref{fig:web_report}) for inspecting the generated content; 
(2) Graph Projection (Figure~\ref{fig:web_graph}) for tracing logical dependencies; 
(3) Question Intervention (Figure~\ref{fig:web_question}) for refining evaluation items; and 
(4) Final Auditing (Figure~\ref{fig:web_eval}) for analyzing performance metrics.

\clearpage



\newpage


\section*{Supplementary material}

# Next-Gen NO-NSAIDs: Viability Challenges Amidst Safety Migration, PPI Paradox, and NaV1.8 Displacement.

**Date:** 2026-01-28

## 1. Abstract

Start of the report. This analysis investigates the structural collapse of the gasotransmitter-releasing non-steroidal anti-inflammatory drug (NSAID) class, specifically Nitric Oxide (NO)-NSAIDs and Hydrogen Sulfide ($H_2S$)-NSAIDs. [global_insight_1][quiz_fact_100][quiz_fact_102] Despite achieving mechanistic validation by resolving the "Three-Hit" model of enteropathy, the class has succumbed to a "Safety Migration" paradox where toxicity is not eliminated but displaced from the gastric mucosa to the liver. [global_insight_1][global_insight_7][global_insight_11] This report argues that the commercial failure of agents such as naproxcinod and otenaproxesul (ATB-346) was not merely a regulatory misfortune but a deterministic outcome of incompatible metabolic trade-offs and an obsolete economic model. [global_insight_1][quiz_fact_98] The emergence of NaV1.8 inhibitors, which bypass the arachidonic acid cascade entirely, has effectively closed the therapeutic window for modified NSAIDs, relegating them to a niche role as genotype-specific rescue therapies. [global_insight_3][global_insight_9][quiz_fact_116]

### 1.1 The Safety Migration Paradox

The central thesis of this investigation is the "Safety Migration" paradox, a phenomenon observed across three generations of modified NSAIDs (benoxaprofen, licofelone, and otenaproxesul). [global_insight_7][global_insight_11] While the addition of gas-releasing moieties (e.g., dithiole-3-thiones) successfully mitigates mucosal injury by maintaining mitochondrial integrity and blood flow, the metabolic processing of these "protective" payloads imposes a significant burden on hepatic function. [global_insight_1][global_insight_7][global_insight_11]

As illustrated in Figure 1, the evolution of the NSAID scaffold reveals a distinct inverse correlation between mucosal safety and hepatic liability. While traditional NSAIDs carry a high risk of gastrointestinal (GI) bleeding with minimal liver toxicity, next-generation agents achieved a profound reduction in ulceration rates—dropping from 42% to 2.5% in Phase 2b trials for otenaproxesul. [global_insight_1][quiz_fact_72][quiz_fact_69][chap_05_fact_21] However, this mucosal victory was counterbalanced by a surge in idiosyncratic drug-induced liver injury (DILI), characterized by transaminase elevations exceeding 5x the Upper Limit of Normal (ULN). [global_insight_1][global_insight_7][chap_05_fact_23] In a post-Vioxx regulatory environment that demands zero tolerance for DILI in non-life-threatening conditions, this "Whac-A-Mole" safety profile renders the class unapprovable for mass-market osteoarthritis indications. [global_insight_1][quiz_fact_71][chap_05_insight_42]

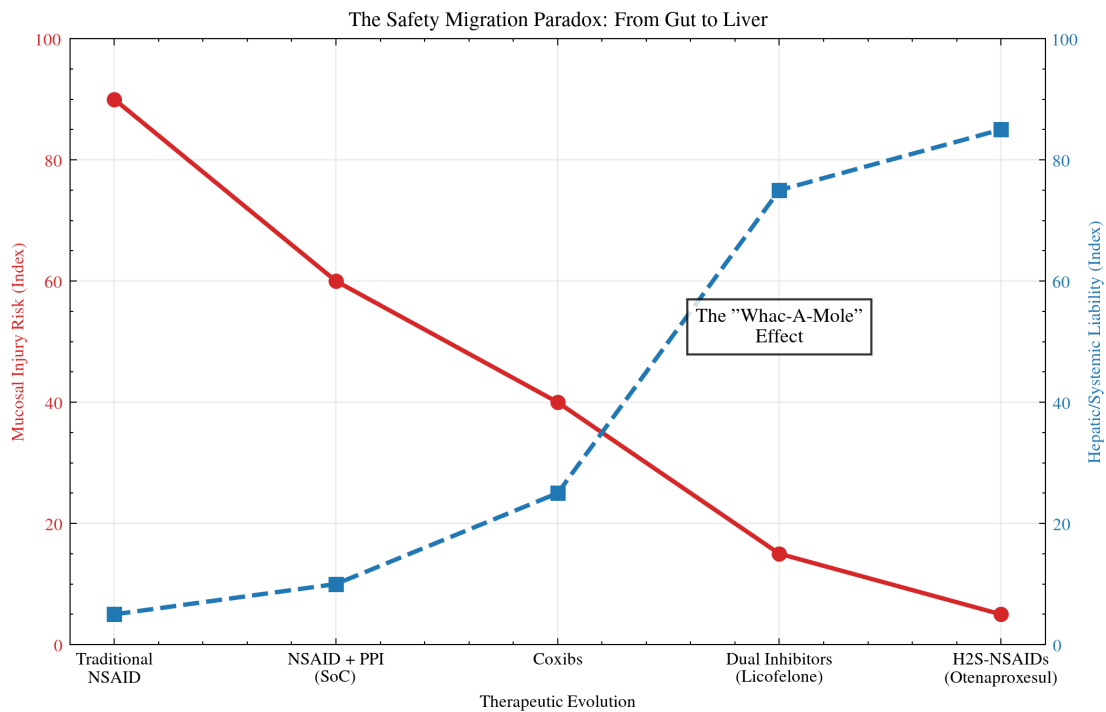

## 1.2 Mechanistic Resolution and the PPI Paradox

Despite the hepatic liabilities, H₂S-NSAIDs demonstrated unequivocal mechanistic superiority in the small bowel, a region where the current Standard of Care (NSAID + PPI) fails. [chap_02_insight_40][quiz_fact_97][quiz_fact_99] The report validates the "PPI Paradox," wherein Proton Pump Inhibitors exacerbate small bowel enteropathy by inducing hypochlorhydria and dysbiosis. [quiz_fact_16][quiz_fact_18][quiz_fact_99][global_insight_8][global_insight_12] This shifts the microbiome toward $\beta$-glucuronidase-producing species, which reactivate NSAID metabolites into toxic aglycones. [chap_01_fact_19][quiz_fact_30][quiz_fact_31][quiz_fact_32]

H₂S-NSAIDs resolve this by serving as inorganic electron donors to the mitochondrial electron transport chain, preventing the bioenergetic collapse that initiates mucosal injury. [chap_03_fact_18][chap_03_residual_information] This restoration of homeostasis is quantified by the Resolution Index ($R_i$), which models the tension between pro-inflammatory drivers and pro-resolving mediators: [chap_01_residual_information]

$$ R_i \propto \frac{[LTB_4] \times [Neutrophil_{influx}]}{[LXA_4] \times [Macrophage_{efferocytosis}]} \tag{1} $$ [chap_01_insight_29][chap_01_residual_information]

Where $[LTB_4]$ represents leukotriene-driven inflammation and $[LXA_4]$ represents Lipoxin-mediated resolution. [quiz_fact_19][chap_01_residual_information] While traditional NSAIDs increase the numerator (via the arachidonic acid shunt), H₂S-NSAIDs and NO-NSAIDs were designed to decrease the numerator (inhibiting neutrophil adherence) and increase the denominator (mimicking Aspirin-Triggered Lipoxins). [chap_01_insight_29][chap_01_residual_information][chap_02_fact_8][chap_02_fact_9][chap_01_fact_14] However, the clinical utility of lowering $R_i$ is nullified if the systemic metabolic cost triggers hepatic necrosis. [global_insight_7][global_insight_11]

## 1.3 Economic Insolvency and the Generic Wall

Beyond biological constraints, the gasotransmitter-NSAID class faces an insurmountable economic barrier defined as the "Generic Wall." [global_insight_2][global_insight_6][global_insight_10][chap_06_insight_16] The cost of preventing NSAID-induced enteropathy via premium agents exceeds the cost of treating the complications they prevent. [global_insight_2][global_insight_6] Our analysis indicates that for the general osteoarthritis population, the "Cost to Prevent" a single hospitalization for small bowel bleeding is approximately $2.75 million, whereas the "Cost of Cure" for such an event is roughly $40,000. [global_insight_6][quiz_fact_105]

This disparity renders the mass-market replacement strategy mathematically impossible. [global_insight_2] To justify the premium pricing ($10–$15/day) required to recoup R&D costs against generic naproxen/omeprazole (<$0.50/day), the therapy would need to target ultra-high-risk cohorts with an annual bleed risk exceeding 10%. [global_insight_2][quiz_fact_106] As detailed in Table 1, the emergence of NaV1.8 inhibitors (e.g., suzetrigine) further erodes this value proposition by offering analgesia without the "Three-Hit" liabilities of COX inhibition, effectively bypassing the need for complex risk stratification. [global_insight_3][global_insight_9][quiz_fact_116]

Table 1: Comparative Viability Matrix of Pain Therapeutics

| Therapeutic Class | Primary Mechanism | GI Safety Profile | Systemic Liability | Economic Viability Status |
|---|---|---|---|---|
| **Standard of Care** (NSAID + PPI) [global_insight_4] | COX Inhibition + Acid Suppression | **Gastric:** High **Small Bowel:** Pathogenic (Dysbiosis) [global_insight_4] [global_insight_8] [global_insight_12] | Renal/CV Risk | **High** (Commoditized <$0.50/day) [global_insight_2] [quiz_fact_106] |
| **NO-NSAIDs** (Naproxcinod) [global_insight_5] | COX Inhibition + NO Release | **Gastric:** Moderate **Small Bowel:** Protective | CV Instability (BP Fluctuations) [quiz_fact_22] [quiz_fact_26] [chap_02_fact_11] [chap_05_fact_13] | **Failed** (Regulatory Rejection) [global_insight_5] [quiz_fact_74] [chap_05_fact_14] |
| **H$_2$S-NSAIDs** (Otenaproxesul) [quiz_fact_28] [quiz_fact_29] | COX Inhibition + H$_2$S Release | **Gastric:** Superior **Small Bowel:** Superior (PPI-Sparing) [quiz_fact_27] [quiz_fact_28] [chap_05_fact_21] [chap_05_insight_38] | Hepatic Toxicity (DILI >5x ULN) [global_insight_1] [quiz_fact_73] [chap_05_fact_23] | **Insolvent** (Bankruptcy/Restructuring) [global_insight_14] [quiz_fact_109] [chap_06_fact_9] |
| **NaV1.8 Inhibitors** (Suzetrigine) [quiz_fact_116] | Voltage-Gated Sodium Channel Blockade | **Gastric:** Neutral (No Risk) **Small Bowel:** Neutral [global_insight_9] | Minimal (Non-Systemic) [global_insight_3] | **High** (Premium Pricing Justified by Safety) [chap_06_fact_10] [quiz_fact_110] |

## 1.4 Conclusion: The Genotype-Specific Beachhead

Consequently, the therapeutic window for "Safer NSAIDs" has structurally closed. [global_insight_3][global_insight_9][global_insight_13] The future of the H$_2$S-NSAID class is restricted to a "Genotype-Specific Rescue" role. [global_insight_4] This applies strictly to the ~30% of the population carrying *CYP2C19* loss-of-function alleles or *FUT2* non-secretor status, for whom

the Standard of Care is mechanistically contraindicated due to accelerated dysbiosis and metabolic failure. [global_insight_4][quiz_fact_112][chap_04_fact_6][chap_04_fact_14][chap_04_fact_15][quiz_fact_43] For the broader market, the paradigm has shifted from "Pathway Modification" (fixing COX toxicity) to "Pathway Bypass" (NaV1.8 inhibition), rendering the complex medicinal chemistry of gasotransmitter-NSAIDs a scientific curiosity rather than a clinical staple. [global_insight_3][global_insight_9][global_insight_13]

## 2. Introduction: The Shift from Gastropathy to Enteropathy

Following the structural collapse of the gasotransmitter-NSAID class described in the previous analysis, it is necessary to dissect the clinical and biological drivers that precipitated this failure. [global_insight_1][global_insight_5][chap_06_insight_15][chap_05_insight_30][chap_05_insight_28] While the commercial insolvency of these agents was sealed by the "Generic Wall" and regulatory intransigence, the scientific impetus for their development remains valid: the "Safety Migration" of NSAID toxicity. [chap_06_insight_16][chap_05_fact_15][global_insight_1][global_insight_7] For three decades, pharmaceutical development focused exclusively on mitigating **gastropathy** (stomach injury), achieving success through Proton Pump Inhibitors (PPIs) and COX-2 selective inhibitors. [chap_05_insight_29][chap_02_insight_27][chap_03_residual_information] However, this upper-GI victory unmasked, and in some cases exacerbated, a more insidious pathology distal to the ligament of Treitz: **NSAID-induced enteropathy**. [chap_05_insight_29][chap_02_insight_27][chap_03_insight_38][global_insight_8]

This chapter defines the shift from an acid-driven model of injury to a multi-factorial cascade involving mitochondrial bioenergetics and microbiome dysbiosis. [chap_03_residual_information][chap_03_insight_45] It establishes the "Three-Hit" pathogenesis model as the scientific benchmark against which next-generation agents like otenaproxesul (ATB-346) and naproxcinod were measured, and ultimately, where the Standard of Care (NSAID+PPI) actively fails. [chap_01_residual_information][chap_03_residual_information][chap_05_insight_32][chap_05_fact_1][global_insight_4]

### 2.1 The 'Three-Hit' Pathogenesis Model

The historical reductionism that equated "GI Safety" with "Stomach Safety" ignored the distinct physiological environment of the small intestine. [chap_02_insight_35] Unlike the stomach, where injury is primarily driven by acid-pepsin digestion of a prostaglandin-depleted mucosa, small bowel injury is a sequential metabolic collapse. [chap_01_residual_information] This is formalized as the **'Three-Hit' Hypothesis**, a cascade that transforms a biochemical interaction into gross tissue necrosis. [chap_01_residual_information][chap_03_residual_information]

1. **Hit 1: Mitochondrial Uncoupling (The Topical Phase):** NSAIDs are lipophilic weak acids. They permeate enterocyte mitochondria and act as protonophores, uncoupling oxidative phosphorylation. This dissipates the proton motive force required for ATP synthesis, leading to the breakdown of intercellular tight junctions (zonula occludens). [chap_03_fact_19][chap_03_fact_20][chap_03_residual_information]

2. **Hit 2: Microvascular Stagnation:** The systemic inhibition of COX-1/2 reduces the synthesis of cytoprotective prostaglandins (PGE2, PGI2), causing vasoconstriction and reducing mucosal blood flow. This ischemia renders the mucosa vulnerable to luminal aggressors. [chap_01_residual_information]




\end{document}